\documentclass{article}

\usepackage[final]{nips_2017}

\usepackage[utf8]{inputenc} 
\usepackage[T1]{fontenc}    
\usepackage{hyperref}       
\usepackage{url}            
\usepackage{booktabs}       
\usepackage{amsfonts}       
\usepackage{nicefrac}       
\usepackage{microtype}      
\usepackage{amsmath}
\usepackage{algpseudocode}
\usepackage{algorithm}
\usepackage{url}
\usepackage{multirow}
\usepackage{graphicx}
\usepackage{color}
\usepackage{mathrsfs}
\usepackage{adjustbox}
\usepackage{setspace}
\usepackage{float}
\usepackage{wrapfig}

\title{Label Denoising Adversarial Network (LDAN) for Inverse Lighting of Face Images}

\author{
  Hao Zhou$^*$ \And Jin Sun$^*$ \And Yaser Yacoob \And David W. Jacobs\\
  University of Maryland, College Park, MD, USA \\
  \texttt{\{hzhou, jinsun, yaser, djacobs\}@cs.umd.edu}\\
}

\begin{document}

\maketitle

\begin{abstract}
Lighting estimation from face images is an important task and has applications in many areas such as image editing, intrinsic image decomposition, and image forgery detection. 
We propose to train a deep Convolutional Neural Network (CNN) to regress lighting parameters from a single face image. 
Lacking massive ground truth lighting labels for face images in the wild, we use an existing method to estimate lighting parameters, which are treated as ground truth with unknown noises. 
%
%
To alleviate the effect of such noises, we utilize the idea of Generative Adversarial Networks (GAN) and propose a Label Denoising Adversarial Network (LDAN) to make use of synthetic data with accurate ground truth to help train a deep CNN  for lighting regression on real face images. 
%
Experiments show that our network outperforms existing methods in producing consistent lighting parameters of different faces under similar lighting conditions.
%
Moreover, our method is 100,000 times faster in execution time than prior optimization-based lighting estimation approaches.
\end{abstract}

\section{Introduction}
\label{sec:introduction}
Estimating lighting sources from an image is a fundamental problem in computer vision.
In general, this is a particularly difficult task when the scene has unknown shape and reflectance properties. 
On the other hand, estimating the lighting of a human face, one of the most popular and well studied objects, is easier due to its approximately known geometry and near Lambertian reflectance.
Lighting estimation can be used in applications such as image editing, 3D structure estimation, and image forgery detection.
This paper focuses on estimating lighting from a single face image.
We consider the most common face image type: near frontal pose.
The same idea can be applied to face images with other poses.

There exist many approaches for lighting estimation from a single face image {\cite{SIRFS,InverseLighting_FG,InverseLighting_ICCV,Forgery17}, however they are not learning-based and rely on complicated optimization during testing, making the process inefficient.
Moreover, the performance of these methods (e.g.,~\cite{SIRFS}) depends on the resolution of face images, and cannot give accurate predictions for low resolution images.

Witnessing the dominant success of neural network models in other computer vision problems such as image classification, we are interested in a supervised learning approach that directly regresses lighting parameters from a single face image. 
Given an input face image, the approach outputs low dimensional Spherical Harmonics coefficients \cite{SH_David,SH_Ravi} of its environment lighting condition. 
%
This is a very difficult problem, especially due to the scarcity of accurate ground truth lighting labels for real face images in the wild.
In fact, building a dataset with realistic images and ground truth lighting parameters is extremely hard and currently there exists no such dataset.

\begin{figure}
\begin{adjustbox}{center}
\begin{tabular}{cc}
\includegraphics[width=5.6cm]{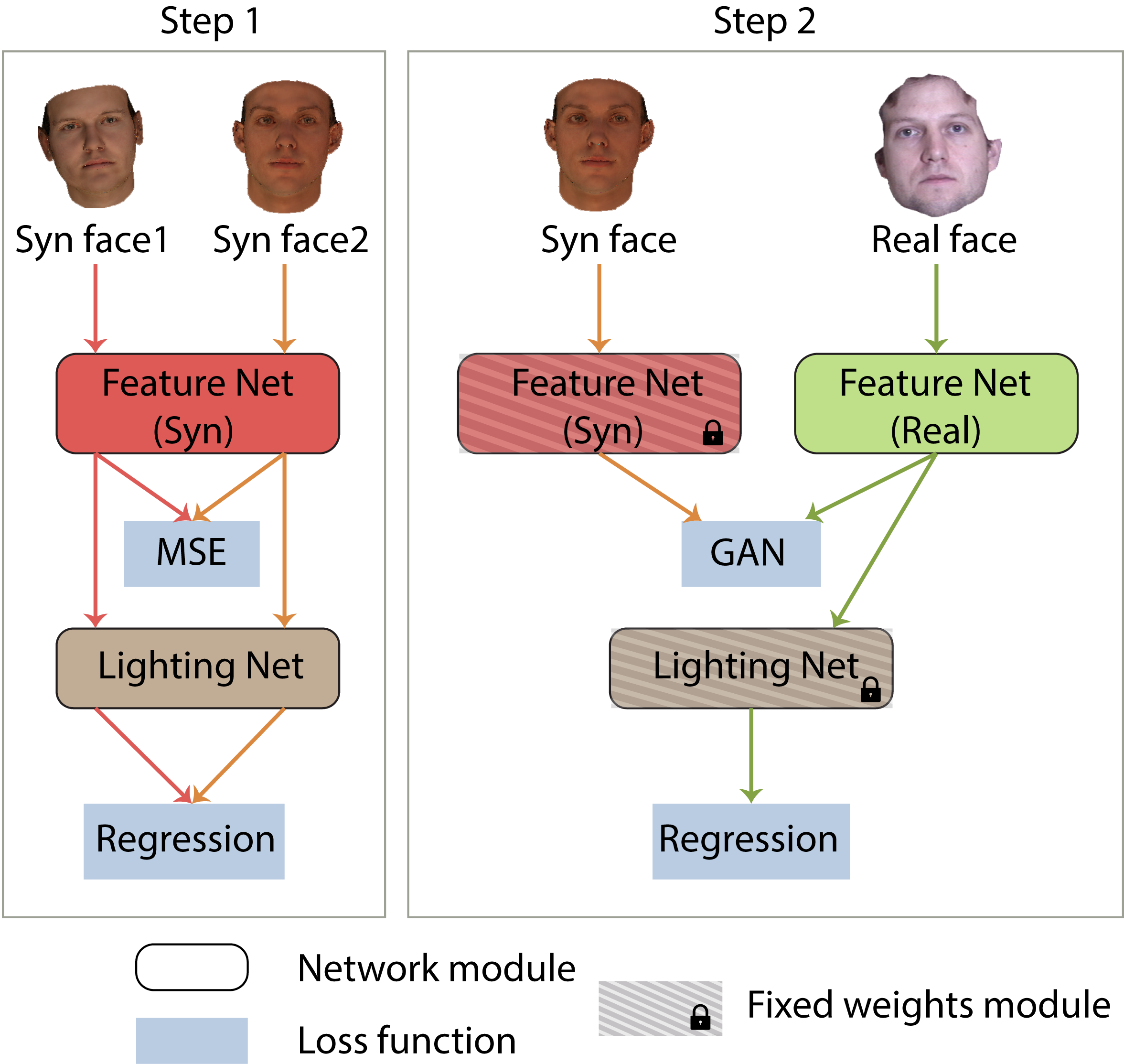}&
\includegraphics[width=6.5cm]{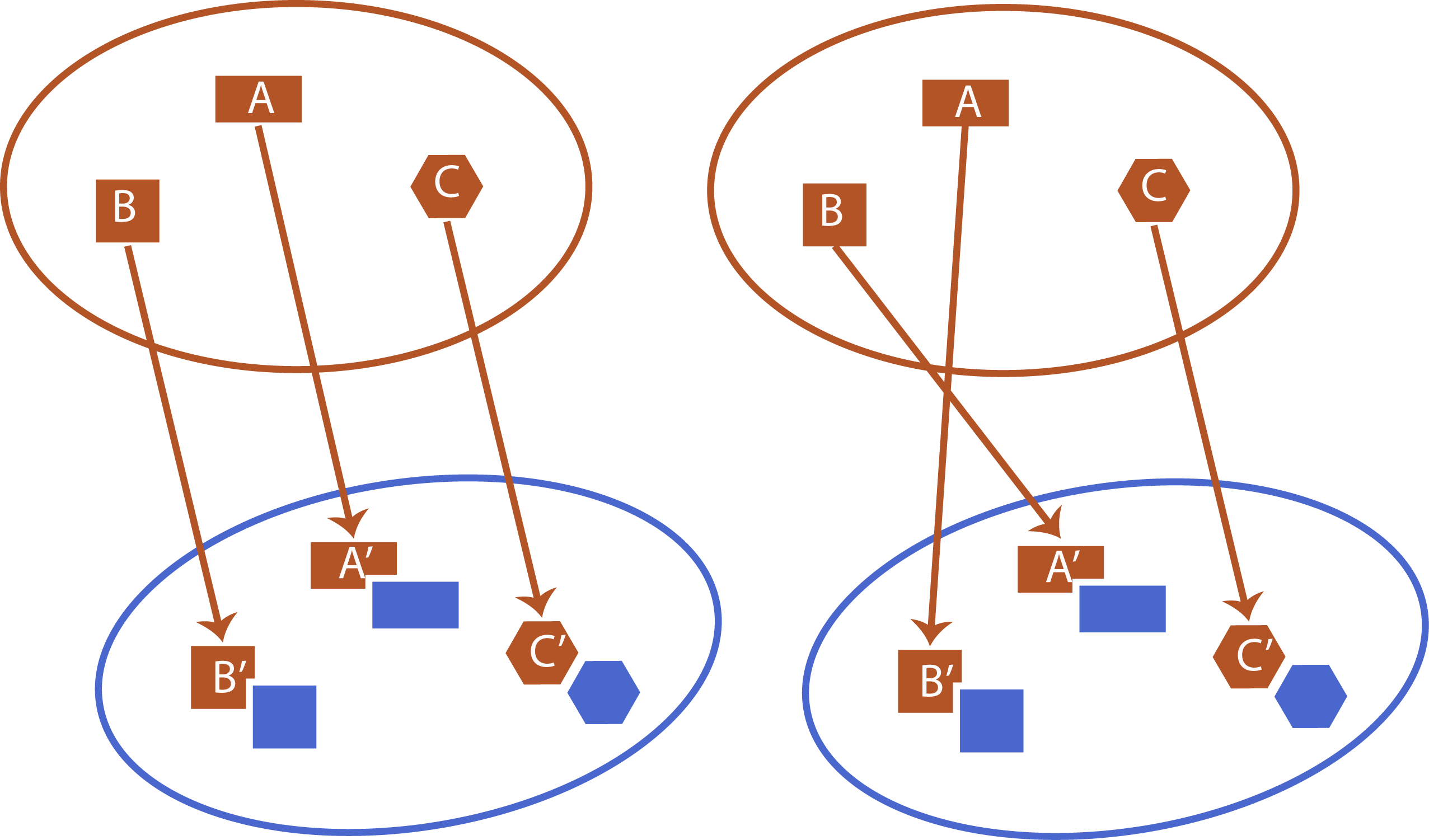}\\
\textbf{(a)} & \textbf{(b)}
\end{tabular}
\end{adjustbox}
\caption{ \textbf{(a)} Training of a LDAN model has two steps: 1) Train the feature net and lighting net for synthetic data; 2) Train the feature net for real data by fixing both the feature net and lighting net trained in step 1. \textbf{(b)} shows two different functions that map data from the source domain to target domain with similar GAN loss. With additional regression loss, our model is encouraged to learn a better behaved mapping function.}
\label{fig:model}
\end{figure}

%
Lacking ground truth labels, we applied an existing method \cite{SIRFS} to estimate lighting parameters of real face images.
However, these lighting parameters are not the real ``ground truth'' as they contain unknown noise.
Synthetic face images, on the other hand, have noise free ground truth lighting labels.
In this work, we show that these synthetic data with accurate labels can help train a deep CNN to regress lighting of real face images: ``denoising'' the unreliable labels.

The proposed method is based on two assumptions: (1) A deep CNN trained with synthetic data is accurate, i.e., it is not affected by any noise; (2) Ground truth labels for real data are noisy, but still contain useful information.
We design the lighting regression deep CNN, which consists of two sub-networks: a feature net that extracts lighting related features and a lighting net that takes these features as input and predicts the Spherical Harmonics parameters.
Based on the first assumption, the lighting net trained with synthetic data is accurate.
However, this lighting net expects lighting related features for synthetic data as input.
To make it work for real data, the lighting related features for real data should be mapped to the same space.
For that purpose, we utilize the idea of Generative Adversarial Networks (GAN) \cite{GAN}. 
Specifically, a discriminator is trained to distinguish between lighting related features from synthetic data and real data, while the feature net (instead of a generator in the standard GAN) is trained to fool the discriminator.
The discriminator and our feature net play a minimax two player game, with the objective of pulling the distribution of lighting related features of real data towards that of the synthetic data. 
Under the second assumption, we have an additional objective of reducing regression loss between predicted lightings and ground truth labels.
Moreover, we design the network to take $64\times 64$ RGB face images so that it will work for low resolution face images.

Figure~\ref{fig:model} (a) illustrates the proposed LDAN model. 
It consists of two steps during training: (1) Train with synthetic data; (2) Fix the feature net for synthetic data and the lighting net, train another feature net for real data with GAN loss and regression loss. 
Eric et al. \cite{DomainGan_Eric} proposed similar ideas for unsupervised domain adaption. 
One difference is that while learning to map target domain to source domain, they only use GAN loss. 
We argue that such mapping can be unexpectedly arbitrary. 
As illustrated by Figure~\ref{fig:model} (b), both mapping $A$ to $A'$, $B$ to $B'$ and mapping $A$ to $B'$ and $B$ to $A'$ make the source and target data have similar distributions. 
This may not be a big issue for classification tasks if $A$ and $B$ belong to the same class. 
However, for regression, this is problematic since every data has its own unique label.
As a result, using the regression loss for real data is critical in our regression problem: it regularize the domain mapping function to have reasonable behavior. 
At the same time, the noises in real data labels are suppressed by training with the GAN loss.

The main contributions of our work are: 1) We are the first to propose a lighting regression network for face images;
2) We propose a novel method: LDAN, to utilize accurate synthetic image lighting labels in training real face images with noisy labels;
3) The proposed method increases the accuracy of \cite{SIRFS} by $9\%$ on quantitative evaluation and is thousands of times faster.

\section{Related Work}
\label{sec:relatedWork}
\textbf{Lighting Estimation from A Single Face Image.}
Estimating lighting conditions from a single face image is an interesting but difficult problem. 
Blanz and Vetter \cite{3DModel} proposed to estimate the ambient and directional light as a byproduct of fitting 3D Morphable Models (3DMM) to a single face image. 
Since then, several 3DMM based methods were proposed \cite{Inverse3D,Forgery17,InverseLighting_ICCV,InverseLighting_FG,SH_Samaras_1}. 
The performance of these methods rely on a good 3DMM of faces. 
However, existing 3DMMs are usually built with face images taken in a controlled environment, so their expressive power (especially the texture model) for faces in the wild is limited \cite{3DWild}. 
Barron and Malik proposed an optimization based method for estimating shape, albedo and lighting for general objects \cite{SIRFS}.
To solve such an underconstrained problem, their method heavily relies on prior knowledge about shape, albedo and lighting of general objects.
Though they achieved promising results, their method is slow and may fail to give a reasonable result for some cases due to the non-convexity of the objective function.
\cite{DC_IGN} proposed to use deep learning to disentangle representations about pose, lighting and identity of a face image. 
The authors only show the effectiveness of their method on synthetic images; whether it could be applied to real face images is still in doubt.
Moreover, their representation of lighting has no physical meaning, making it difficult to use in other applications.

\textbf{Learning with Noisy Labels.}
Learning with noisy labels has attracted the interest of researchers for a long time.
\cite{NoiseIntro} gives a comprehensive introduction to this problem. 
With the development of deep learning, many research studies have now focused on how to train deep neural networks with noisy labels \cite{NoiseLabel_Hinton,NoisyLabel_cnn,NoiseLabel_ICLR2016,NoiseLabel_cvpr2015,NoiseLabel_correct,NoiseLabel_ICDM2016}. 
\cite{NoiseLabel_Hinton,NoisyLabel_cnn,NoiseLabel_correct,NoiseLabel_ICDM2016} assume the probability of a noisy label only depends on the noise-free label but not on the input data, and try to model the conditional probability explicitly.
\cite{NoiseLabel_cvpr2015} models the type of noise as a hidden variable and proposes a novel probabilistic model to infer the true labels.
\cite{NoiseLabel_ICLR2016} proposed to use CNNs pre-trained with noise-free data to help select data with noisy labels in order to better handle the noise. 
All the above mentioned methods focus on classification problems and a considerable portion of the data are assumed to have noise-free labels. 
However, estimating lighting from face images is a regression problem, and the translation probability from noise-free label to noisy label is much more difficult to model.
Moreover, almost all the labels of our data are noisy.
As a result, we are dealing with a much harder problem than the methods mentioned above.

\textbf{GAN for Domain Adaption.}
Since Goodfellow et al. \cite{GAN} first proposed Generative Adversarial Networks, several works have been using this idea for unsupervised Domain Adaption \cite{DomainGan_ICML2015,DomainGan_Eric,DomainGan_Swami,DomainGan_Kuniaki}.
%
%
All these methods solve a problem in which the labels in the target domain are not enough to train a deep neural network.
However, the problem we try to solve is intrinsically different from theirs in that the labels in the target domain are sufficient, but all these labels are noisy. 
Moreover, all these methods apply domain adaption to classification tasks where GAN loss is enough to achieve a good performance. 
On the contrary, GAN loss alone cannot work in our regression task.
Though GAN loss could map the distribution of data in the target domain to that of the source domain, for a single point in the target domain, the mapping is arbitrary which is problematic as every data point has its unique label in a regression task.

\section{Proposed Method}
\label{sec:proposedMethod}
\subsection{Spherical Harmonics}
Existing methods \cite{SH_David,SH_Ravi} have shown that for convex objects with Lambertian reflectance and distant light sources, the lighting of the environment can be well estimated by $9$ (gray scale) or $27$ (color) dimensions of Spherical Harmonics (SH).
In this paper, we use SH as the lighting representation as it has been widely used to represent the environmental lighting in face related applications as suggested in \cite{SH_PS,SH_Samaras_1,SH_Samaras_2,SIRFS,Forgery07,Forgery17}.

All dimensions of SH can be fully recovered from an image if the pixels are equally distributed over a sphere. 
%
%
However, the pixels of a face image, loosely speaking, are distributed over a  hemisphere.
The SH that can be recovered from a face image, as discussed in \cite{PCALight}, lie in a lower dimensional subspace, and the SH for faces under different poses lie in different subspaces. 
As a result, we consider regressing the SH in a lower dimensional subspace instead of the original $27$ dimensional SH and focus on near frontal faces since most face images are taken under this pose.
%

Taking the red color channel as an example, we now show how to get the lower dimensional subspace of SH for near frontal faces.
Let $\mathbf{I}_r$ be a column vector: each element represents one pixel value of a face image for the red channel, then $\mathbf{I}_r = \Lambda_r Y\mathbf{l}_r$.
$\Lambda_r$ is a $n\times n$ diagonal matrix, each element of which is the albedo of the corresponding pixel, $\mathbf{l}_r$ is a 9 dimensional SH parameters vector, $Y$ is a $n \times 9$ matrix and $n$ is the number of pixels in the image.
Each column of $Y$ corresponds to one SH base image whose elements are determined by the normal of the corresponding pixel (see \cite{SH_David}).
By applying SVD on $Y$, we get $Y = UDV^T$, then $\mathbf{I}_r = \Lambda_r UDV^T\mathbf{l}_r$.
$V$ is a $9 \times 9$ matrix that spans the entire $9$ dimensions of SH.
We use synthetic data to get $V$ since we know the ground truth normal of every pixel and thus $Y$ is known.
We then only keep the $6$ columns of $V$, denoted as $V_6$, corresponding to the largest $6$ singular values since they capture $99\%$ energy of the singular values. 
With $V_6$, we project all the SH to their $18$ dimensional subspace throughout the experiments.

\subsection{Label Denoising Adversarial Network}
Training a regression deep CNN needs a lot of data with ground truth labels. 
However, getting the ground truth lighting parameters from a realistic face image is extremely difficult. 
It usually needs a mirror ball or panorama camera which is carefully set up to record an environment map relative to the position of the face.
Thus, it is very difficult to get enough data with ground truth labels to train a deep CNN to regress the lighting of a face.
Instead, we adapted \cite{SIRFS} to predict lighting parameters from a large number of face images. 
These parameters are then projected to a lower dimensional subspace using $V_6$ discussed above.
We use these projected lighting parameters as noisy ground truth labels and denote them as $\mathbf{\hat{y}}_r$.
Together with real face images $\mathbf{r}$, ($\mathbf{r}$, $\mathbf{\hat{y}}_r$) will be used as (data, label) pair to train a deep CNN to regress lighting parameters.
One problem with these ground truth labels is that they are noisy; directly training a deep CNN with these data cannot give the best performance.
%

We propose to use synthetic face images whose ground truth lighting parameters are known to help train a deep CNN.
The proposed deep CNN has two sub-networks: a feature network that is used to extract lighting related features; a lighting network that takes lighting related features as input and predicts SH for the face images.
For synthetic data $\mathbf{s}$, we denote its feature network as $\mathcal{S}$ and its lighting network as $\mathcal{L}$.
Then the predicted SH is represented as $\mathbf{y}_s = \mathcal{L}(\mathcal{S}(\mathbf{s}))$.
Since $\mathcal{S}$ and $\mathcal{L}$ are trained using synthetic data whose ground truth labels are known, they are accurate. 
Feature network $\mathcal{R}$ and Lighting network $\mathcal{L}_r$ for real data, on the other hand, will both be affected by the noisy labels if directly trained using the noisy ground truth of real data.
To alleviate the effect of noisy labels, we propose to use $\mathcal{L}$ as the lighting net for real data, i.e., $\mathcal{L}_r = \mathcal{L}$, since it is not affected by noise.
However, since $\mathcal{L}$ is trained using synthetic data, it only works if the input is from the space of lighting related features of synthetic data.
As a result, $\mathcal{R}$ needs to be trained such that the lighting related features for real data will be mapped into the same space as synthetic data.

Given a set of synthetic images $\mathbf{s}$ and their ground truth labels $\mathbf{y}^*_s$, we train feature net $\mathcal{S}$ and lighting net $\mathcal{L}$ through the following loss function:
\begin{eqnarray}
\min_{\mathcal{S}, \mathcal{L}}\sum_{(i,j)\in \Omega}\underbrace{\nu[(\mathcal{L}(\mathcal{S}(\mathbf{s}_{i})) - \mathbf{y}_{si}^{*})^2 + (\mathcal{L}(\mathcal{S}(\mathbf{s}_{j})) - \mathbf{y}_{si}^{*})^2]}_\text{regression loss for synthetic} + \underbrace{\lambda(\mathcal{S}(\mathbf{s}_{i}) - \mathcal{S}(\mathbf{s}_{j}))^2}_\text{feature loss}, \label{eq:synthetic}
\end{eqnarray}
where $\mathbf{s}_{i}$ and $\mathbf{s}_{j}$ are a pair of synthetic images with the same SH lighting, different IDs, and different small random deviations from frontal pose. 
$\mathbf{y}_{si}^{*}$ represents their ground truth label.
$\Omega$ is a set containing all such pairs.
$\nu$ and $\lambda$ are weight coefficients. 
Besides the regression loss, we also add a MSE feature loss that enforces the lighting related features of face images with the same SH to be the same. 
This encourages the lighting related features to contain no information about face ID and pose.

With trained $\mathcal{S}$ and $\mathcal{L}$, we need to train the feature net $\mathcal{R}$ for real face images $\mathbf{r}$ so that the lighting related features for real data ($\mathbf{f}_r = \mathcal{R}(\mathbf{r})$) lie in the same space as that of synthetic data ($\mathbf{f}_s = \mathcal{S}(\mathbf{s})$).
Our idea is inspired by recently proposed Generative Adversarial Networks (GAN) \cite{GAN}, which have proved to be very effective to synthesize realistic images.
In our setting, a discriminator $\mathcal{D}$ is trained to distinguish $\mathbf{f}_r$ and $\mathbf{f}_s$, while $\mathcal{R}$ is trained so that $\mathbf{f}_r$ would make $\mathcal{D}$ fail. 
By playing this minmax game, the distribution of $\mathbf{f}_r$ will be close to that of $\mathbf{f}_s$.
Wasserstein GAN (WGAN) \cite{WGAN} is used as our training strategy since it can alleviate the ``mode dropping'' problem and generate more realistic samples for image synthesis.
However, making the distribution of $\mathbf{f}_r$ and that of $\mathbf{f}_s$ similar is not enough for our regression problem since the mapping can be arbitrary to some extent.
As shown in Figure~\ref{fig:model} (b), both these two mappings would make two sets of points have similar distributions, but not both of them are correct since every point has its unique label. 
Based on the assumption that noisy ground truth of real data is reasonably close to the ground truth data, we use them as ``anchor points'' during training.
As a result, the final loss function for our problem is defined as follows:
\begin{eqnarray}
\min_{\mathcal{R}, \mathcal{S}, \mathcal{L}}\max_{\mathcal{D}} 
& \underbrace{\mathbb{E}_{\mathcal{S}(\mathbf{s})\sim \mathbb{P}_s }[\mathcal{D}(\mathcal{S}(\mathbf{s})] - \mathbb{E}_{\mathcal{R}(\mathbf{r})\sim \mathbb{P}_r }[\mathcal{D}(\mathcal{R}(\mathbf{r}))]}_\text{GAN loss} 
+ \underbrace{\mu \sum_i(\mathcal{L}(\mathcal{R}(\mathbf{r}_{i})) - \mathbf{\hat{y}}_{ri})^2}_\text{regression loss for real} \nonumber \\
&+ \sum_{(i,j)\in \Omega}(
\underbrace{\nu [(\mathcal{L}(\mathcal{S}(\mathbf{s}_{i})) - \mathbf{y}_{si}^{*})^2 + (\mathcal{L}(\mathcal{S}(\mathbf{s}_{j})) - \mathbf{y}_{si}^{*})^2]}_\text{regression loss for synthetic} + 
\underbrace{\lambda(\mathcal{S}(\mathbf{s}_{i}) - \mathcal{S}(\mathbf{s}_{j}))^2}_\text{feature loss}) \label{eq:wgan}
\end{eqnarray}
where $\mathbb{P}_s$ and $\mathbb{P}_r$ are the distributions of lighting related features for synthetic and real images respectively.

Following \cite{GAN,WGAN}, the discriminator $\mathcal{D}$ and feature net $\mathcal{R}$ are trained alternatively.
While training $\mathcal{D}$, RMSProp \cite{RMSProp} is applied and Adadelta \cite{Adadelta} is used to train $\mathcal{S}$, $\mathcal{R}$ and $\mathcal{L}$  as discussed in \cite{WGAN}.
The details on how to train the whole model are illustrated in Algorithm~\ref{al:CNN_A}.

\begin{algorithm}
\caption{Training procedure for LDAN}
\label{al:CNN_A}
\begin{algorithmic}[1]
\State {Train $\mathcal{S}$ and $\mathcal{L}$ for synthetic data using loss function in Equation~\ref{eq:synthetic} by Adadelta.} 
\State {Compute $\mathbf{f}_s = \mathcal{S}(\mathbf{x}_s)$ for all synthetic images.}
\For {number of training iterations}
	\For {$k$= 1 to 5 epochs}
    	\State{Train discriminator $\mathcal{D}$ through the following loss using RMSProp:
        \begin{eqnarray}
        	\max_{\mathcal{D}}\mathbb{E}_{\mathbf{f}_s\sim \mathbb{P}_s }[\mathcal{D}(\mathbf{f}_s)] - \mathbb{E}_{\mathcal{R}(\mathbf{r})\sim \mathbb{P}_r}[\mathcal{D}(\mathcal{R}(\mathbf{r}))] \nonumber
        \end{eqnarray}} 
    \EndFor
    \For {$k$=1 to 2 epochs}
    	\State {Train $\mathcal{R}$ thought the following loss using Adadelta:
        \begin{equation}
        	\min_{\mathcal{R}} - \mathbb{E}_{\mathcal{R}(\mathbf{r})\sim \mathbb{P}_r }[\mathcal{D}(\mathcal{R}(\mathbf{r}))] + \mu \sum_i(\mathcal{L}(\mathcal{R}(\mathbf{r}_{i})) - \mathbf{\hat{y}}_{ri})^2 \nonumber
        \end{equation}}
    \EndFor
\EndFor
\end{algorithmic}
\end{algorithm}


\section{Experiments}
\label{sec:experiments}

\subsection{Data Collection}
\textbf{Real Face Images:} The proposed LDAN requires a large number of both synthetic and real face images for training. 
To collect the real face images, we download images with faces from the Internet. 
The SIRFS method proposed by Barron and Malik \cite{SIRFS} is then applied to these face images to get the noisy ground truth of SH for lighting. 
Since SIRFS was proposed to estimate lighting for general objects, the prior they use is not face-specific.
To get a better constraint for a face shape, we apply Discriminative Response Map Fitting (DRMF) \cite{Y1} to estimate the facial landmarks and pose.
Then, a 3DMM \cite{3DModel} is fitted to get an estimation of the face depth map which is used as a prior to constrain the face shape estimation of SIRFS. 
We collected $40,000$ faces with noisy ground truth SH for training. 

\textbf{Synthetic Face Images:} We apply the 3D face model provided by \cite{BaselFace} to generate $40,000$ pairs of faces.
Each pair of these faces are under the same lighting but with different identities and a small random variation with respect to frontal pose. 

\textbf{MultiPie:} The MultiPie dataset \cite{MultiPie} contains a large number of face images of different IDs taken under different poses and illumination conditions. 
From this data set, $4,973$ face images are chosen, which contain $250$ IDs in frontal pose under $19$ lighting conditions. 
Though the ground truth lighting parameters are not provided for each of these face images, the lighting condition group under which a face image is taken is given. 
This data is used only for evaluation in our experiments.

\subsection{Implementation Details}
We use the same network structure for feature net $\mathcal{S}$ and $\mathcal{R}$. 
We apply the ResNet structure \cite{ResNet} to define a feature net.
It takes a $64\times 64$ RGB face image as input and outputs a $128$ dimensional feature vector.
We use several fully connected layers to define our lighting net $\mathcal{L}$ and discriminator $\mathcal{D}$.
The lighting net outputs $18$ dimensional lighting parameters and $\mathcal{D}$ outputs the score for being a lighting related feature of real data.
Please refer to the supplementary material for details on the network structures.

While training the proposed model, we first train discriminator $\mathcal{D}$ for $5$ epochs and then train feature net $\mathcal{R}$ for $2$ epochs.
We notice that in this way, $\mathcal{D}$ will be fully trained to distinguish $\mathbf{f}_s$ and $\mathbf{f}_r$ and $\mathcal{R}$ will be trained so that $\mathcal{D}$ will fail. 
We alternate these two steps for $50$ iterations.
We choose $\mu=50$, $\nu=50$ and $\lambda=0.5$ in Equation~(\ref{eq:wgan}) so that different losses are roughly balanced.
Our algorithm is implemented using Keras \cite{keras} with Tensorflow \cite{tensorflow} as backend.

\subsection{Evaluation Metric}
Since ground truth lighting parameters for real face images are not available, it is difficult to evaluate the accuracy of regressed lighting quantitatively. 
We propose an ``indirect'' quantitative evaluation metric based on classification, and test our method on the MultiPie data set, which contains face images taken under $19$ lighting conditions.
More specifically, after regressing the SH for each test face image, $90\%$ of them are used to compute the mean SH for each lighting condition.
Then, each of the rest of the face images are assigned to the $19$ lighting conditions based on the Euclidean distance between its estimated SH and the mean SH.
We carry out $10$ cross validations for this classification measurement to make use of all the data.
%
%


\subsection{Experimental Results}
To show the effectiveness of the proposed method, we compare our results with the SIRFS \cite{SIRFS} based method in this section.
In SIRFS, the shading of a face is formulated in logarithm space, i.e. $\log \{s_i\}= Y_i \mathbf{l}$ where $s_i$ is the shading at the $i$-th pixel, $Y_i$ is the $i$-th row of $Y$ and $\mathbf{l}$ represents the SH. 
$\mathbf{l}$ estimated in this way is not the correct SH. 
To estimate the correct SH lighting, we assume that the normal of each pixel estimated by SIRFS is in Euclidean space instead of logarithm space. 
This assumption is reasonable since we adapted the SIRFS method by estimating a face depth map using 3DMM in Euclidean space, and constrained the estimated face shape to be consistent with it.
Supposing $\hat{\mathbf{l}}$ is the correct SH, the shading can be found by $s_i = Y_i\hat{\mathbf{l}}$.
Then $\hat{\mathbf{l}}$ could be achieved by solving the following equation:
\begin{eqnarray}
Y\hat{\mathbf{l}} = \exp{\{Y\mathbf{l}\}}. \label{eq:correction}
\end{eqnarray}
This is an over complete linear equation as the number of pixels is larger than the dimension of SH.

\begin{table}
\caption{Accuracy of different methods. Standard deviation is shown in the bracket for learning based methods.}
\label{Tb:classification}
\begin{adjustbox}{center}
\begin{tabular}{ccccccc}
       & SIRFS log  & SIRFS SH   & REAL     &  LDAN  & Model B   & Model C   \\
\hline
top-1 ($\%$) & $60.72$ & $56.04$ & $61.29$ $(\pm 1.8)$ & $\mathbf{65.87}$ $(\pm 1.2)$ & 56.71 $(\pm 1.9)$ & 64.26 $(\pm 1.5)$\\ 
top-2 ($\%$) & $79.65$ & $74.39$ & $81.95$ $(\pm 1.3)$ & $\mathbf{85.17}$ $(\pm 1.1)$ & 77.77 $(\pm 1.6)$ & 84.37 $(\pm 1.2)$\\
top-3 ($\%$) & $87.27$ & $83.74$ & $90.59$ $(\pm 0.7)$ & $\mathbf{92.46}$ $(\pm 0.5)$ & 87.75 $(\pm 1.1)$ & 92.12  $(\pm 0.7)$
\end{tabular}
\end{adjustbox}
\end{table}

Table~\ref{Tb:classification} compares the proposed method with the SIRFS based method using the classification measurement on the MultiPie data set. 
We denote the original output of SIRFS method as SIRFS log, and SIRFS SH is used to denote the corrected SH using Equation~(\ref{eq:correction}). 
We test these two methods on the original resolution of the MultiPie data which is roughly $220\times 270$ after cropping the faces.
Note that for a face image of size $64\times 64$ which is the input size for LDAN, the 3DMM we use cannot predict accurate face shapes, resulting in inaccurate estimation by SIRFS.
REAL in Table~\ref{Tb:classification} represents our baseline method which uses SIRFS SH as ground truth to train a deep CNN without synthetic data. 
REAL and LDAN are trained $5$ times and the mean accuracies are shown in Table~\ref{Tb:classification}.
We notice that SIRFS SH, which solves Equation~(\ref{eq:correction}) based on SIRFS log, performs worse than SIRFS log. 
According to Equation~(\ref{eq:correction}), the accuracy of SIRFS SH depends not only on the accuracy of SIRFS log, but also on the accuracy of estimated normals. 
The noisy estimation of normals would make the performance of estimated SIRFS SH more noisy.
The performance of REAL is better than SIRFS SH, though it is trained directly using the output of SIRFS SH as the ground truth label.
This shows that by observing a large amount of data, the deep CNN itself can be robust to noise to some extent. 
This is an advantage for learning based methods compared with optimization based algorithms.
LDAN outperforms the REAL by more than $4\%$ for top-1 accuracy, showing the effectiveness of the proposed method.

\begin{figure}
\centering
\includegraphics[width=8cm]{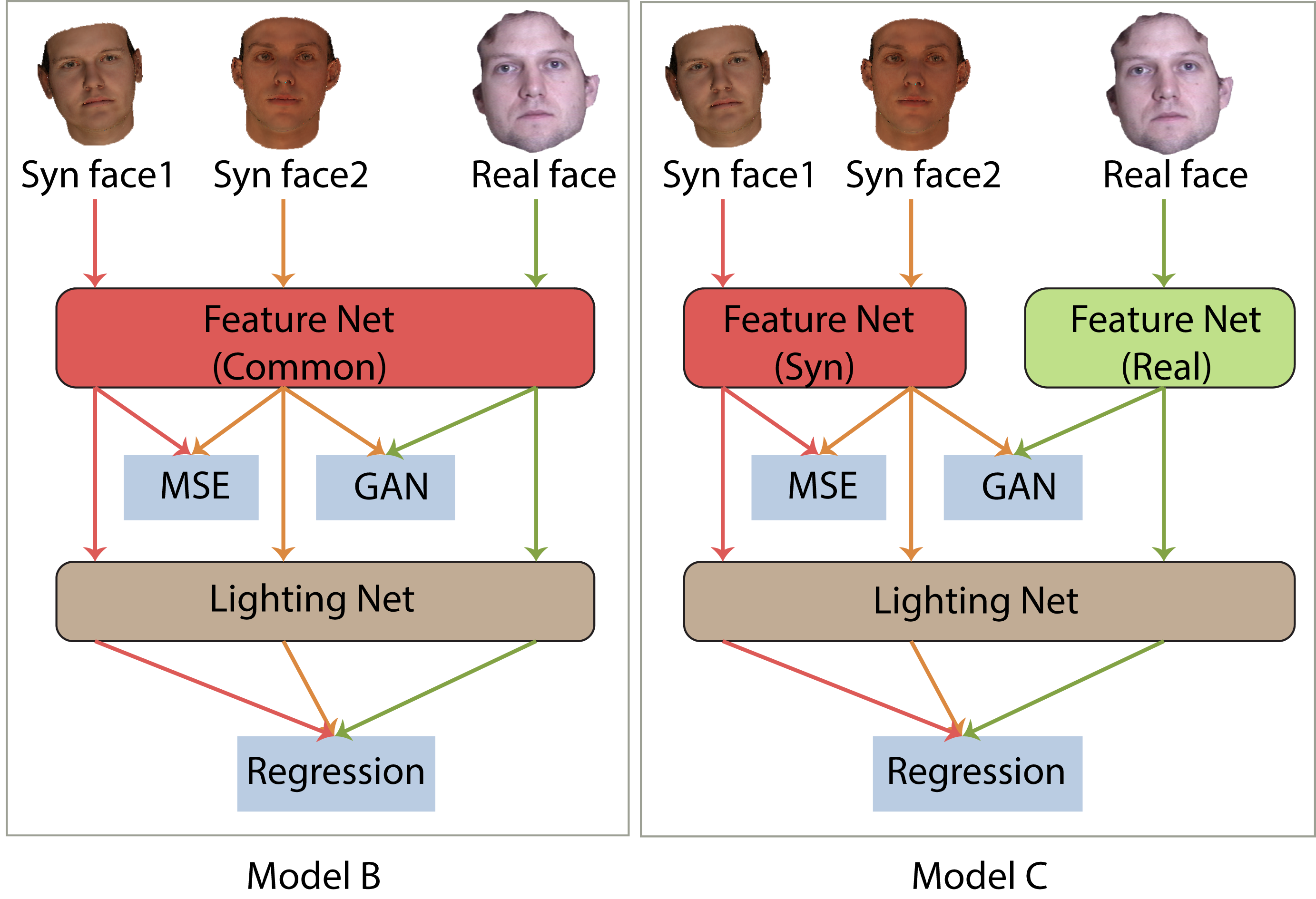}
\caption{Two models we use to compare with the proposed LDAN.}
\label{fig:Model_AB}
\end{figure}
We further propose two other models to compare with LDAN as shown in Figure~\ref{fig:Model_AB}.
Different from LDAN, Model B and Model C learn the feature nets for synthetic and real data simultaneously and map the lighting related features of them to the same space \footnote{Please refer to the supplementary material for the details of how to train Model B and Model C.}.
These two models are inspired by \cite{DomainGan_ICML2015} and \cite{DomainGan_Kuniaki}.
For Model B, synthetic and real data share the same feature net.
Since synthetic data and real data are quite different from each other, using a single feature net is difficult to make their lighting features have the same distribution, and we do not expect good performance.
Model C defines different feature nets for synthetic and real data.
The difference between Model C and LDAN is that Model C tries to map lighting related features for synthetic and real data to the same space  which might be different from that learned with synthetic data alone, whereas LDAN tries to directly map lighting related features of real data to the space of synthetic data.
Intuitively, compared with LDAN, Model C is more easily affected by the noisy labels of real data since the training of the feature net for synthetic data is affected by the real data.
Model B and C are also trained $5$ times and their mean accuracies are shown in Table~\ref{Tb:classification} for comparison.
We notice that Model B performs even worse than REAL, which shows that a single feature net for both synthetic and real data is not enough.
LDAN and Model C outperform all other methods in Table~\ref{Tb:classification}. 
%
%
Moreover, LDAN performs slightly better than Model C, showing that it is more robust to the noise in the ground truth of real data.

\begin{figure}
\begin{adjustbox}{center}
\begin{tabular}{cc}
\includegraphics[width=4cm]{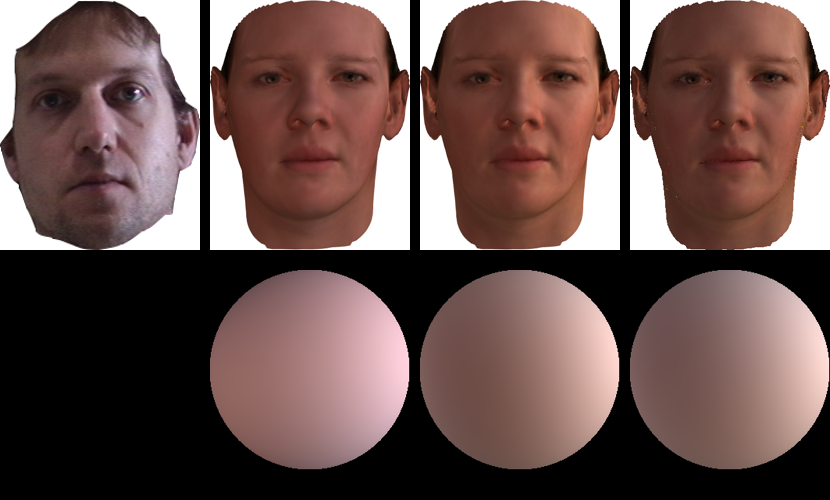} & 
\includegraphics[width=4cm]{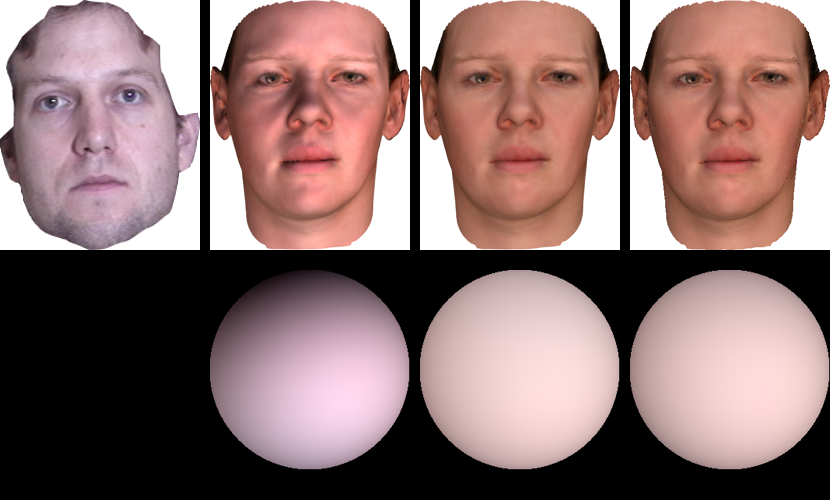} \\
(a) & (b)
\end{tabular}
\end{adjustbox}
\caption{The images in the first row of (a) (b) are: MultiPie face image, synthetic face image rendered using lighting estimated by SIRFS SH, REAL and LDAN from the MultiPie face image. The second row shows a hemisphere rendered by the corresponding lighting.}
\label{fig:example}
\end{figure}

\begin{table}
\caption{Results of ablation study. Standard derivation is shown in the bracket.}
\label{Tb:ablation}
\begin{adjustbox}{center}
\begin{tabular}{cccc}
       & LDAN  & LDAN w/o GAN   & LDAN w/o Regression  \\
\hline
top-1 ($\%$) & $\mathbf{65.87}$ $(\pm 1.2)$ & 60.58 $(\pm 0.5)$ & 30.71 $(\pm 4.1)$\\ 
top-2 ($\%$) & $\mathbf{85.17}$ $(\pm 1.1)$ & 81.88 $(\pm 0.5)$ & 48.47 $(\pm 5.7)$\\
top-3 ($\%$) & $\mathbf{92.46}$ $(\pm 0.5)$ & 90.75 $(\pm 0.6)$ & 60.96 $(\pm 6.4)$
\end{tabular}
\end{adjustbox}
\end{table}

To investigate the effectiveness of GAN loss and regression loss, we carry out ablation studies for LDAN. 
We train the feature net for real data without GAN loss and regression loss for $5$ times respectively and compare the results with LDAN in Table~\ref{Tb:ablation}.
Without GAN loss, the performance of LDAN is similar to REAL in Table~\ref{Tb:classification}, which means that synthetic data could not help to train a better deep CNN for regressing lighting in this case.
Without regression loss, on the other hand, the performance of LDAN drops dramatically.
This is because the way of mapping the distribution of lighting related features of real data to that of synthetic data is arbitrary as shown in Figure~\ref{fig:model} (b).
This is problematic for a regression task where each data has its unique label.
Having noisy ground truth as ``anchor points'', as we do in LDAN, can alleviate this problem and give much better results.

Figure~\ref{fig:example} shows some face images synthesized using the SH parameters estimated by SIRFS SH, REAL and LDAN from MultiPie images. 
For Figure~\ref{fig:example} (b), the lighting of the environment distributes uniformly based on the MultiPie face image, however, SIRFS SH predicts the lighting comes from the lower right part. 
We notice that face images synthesized using lighting estimated by REAL and LDAN are visually similar, but LDAN predicts more consistent lighting for face images taken under the same lighting condition as shown by our classification results.



\begin{figure}
\centering 
\begin{tabular}{cc}
\includegraphics[width=5cm]{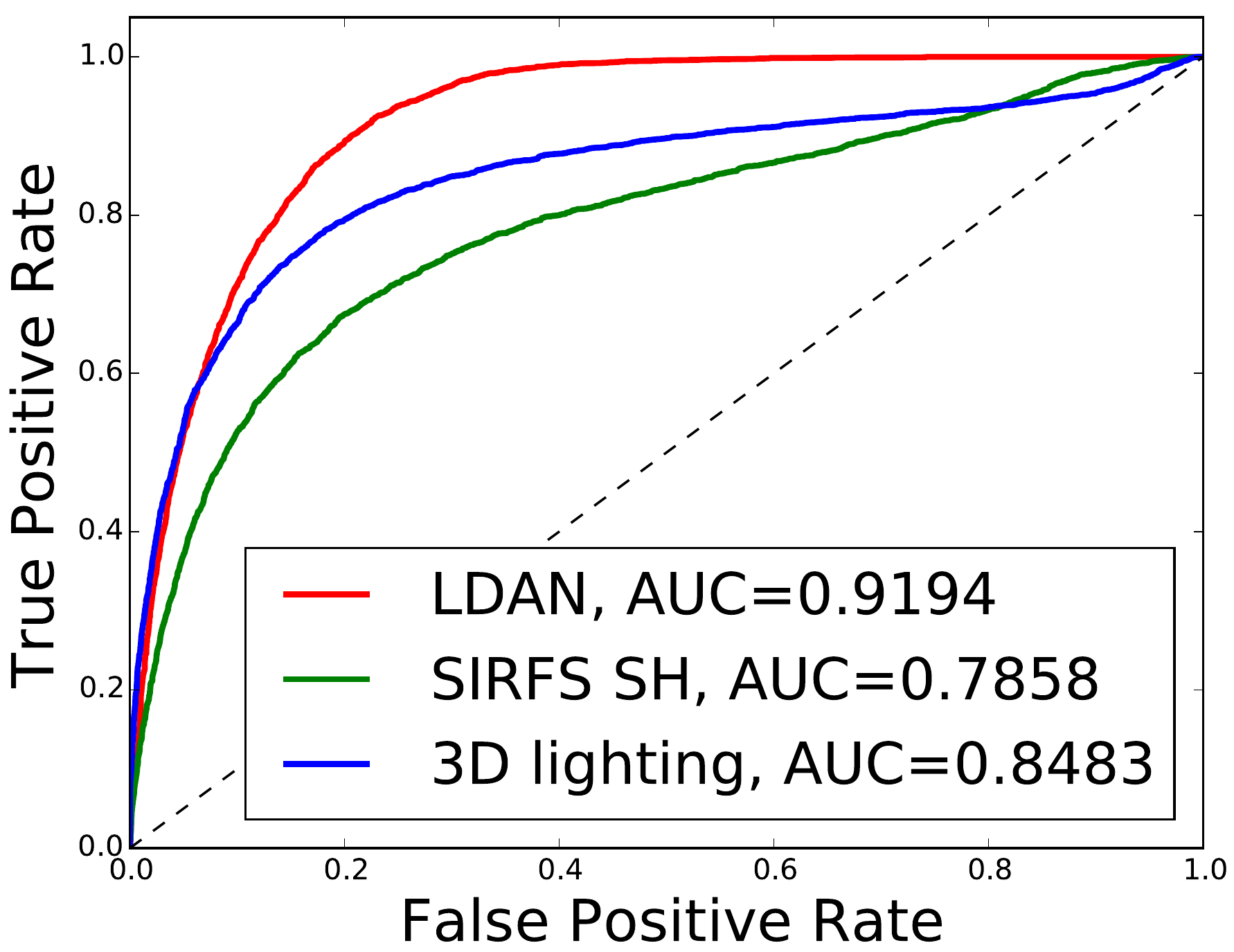} &
\includegraphics[width=5cm]{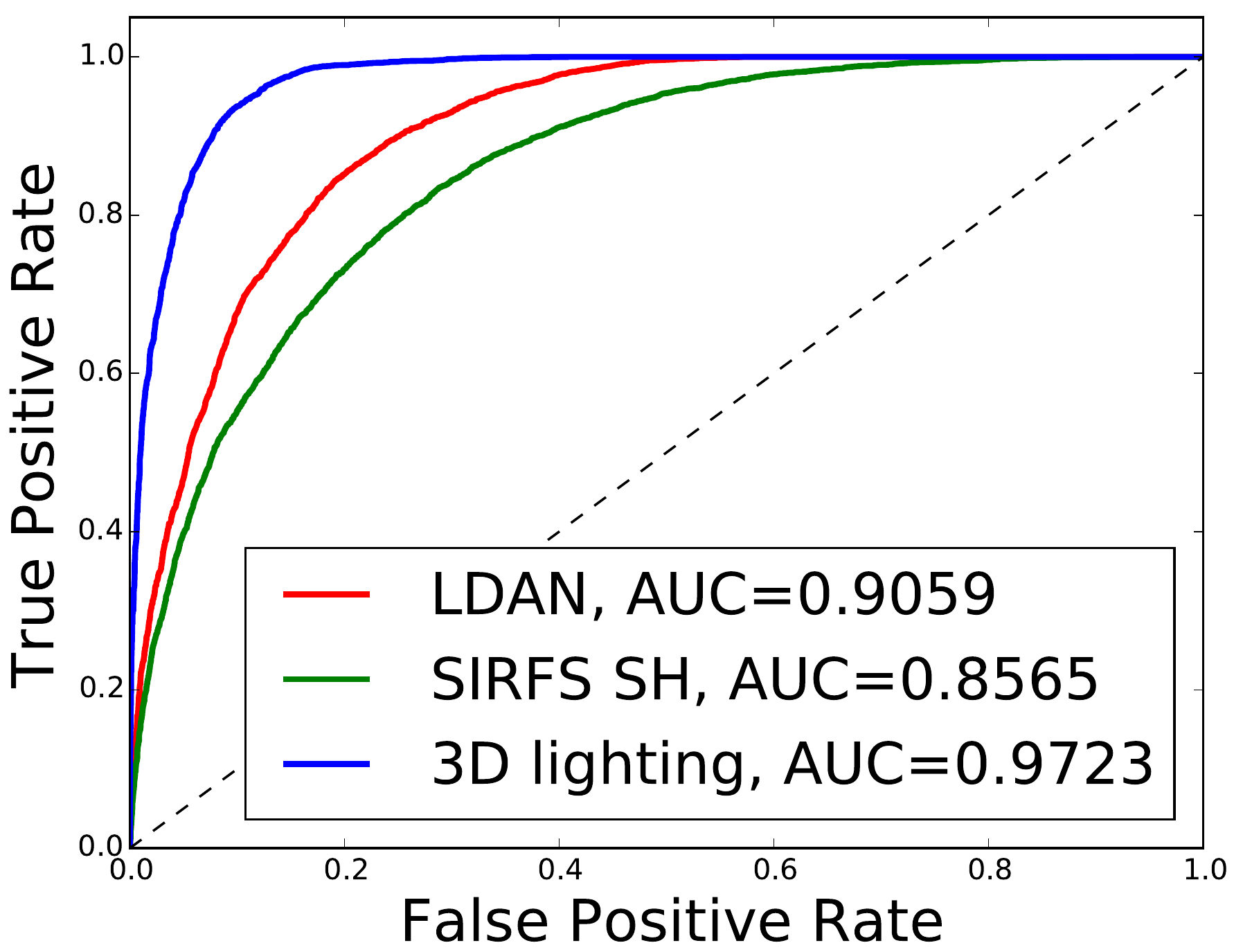} \\
\textbf{(a)} Euclidean distance & \textbf{(b)} $Q$-measure
\end{tabular}
\caption{\textbf{(a)} and \textbf{(b)} compare proposed model with \cite{Forgery17} under Euclidean distance and $Q$-measure respectively.}
\label{fig:ROC}
\end{figure}

Peng et al. \cite{Forgery17} proposed to estimate SH from face images for the application of image forgery detection and achieved start-of-the-art results. 
In experiments of MultiPie, they use two face images, one frontal and one profile, to estimate accurate face normals in order to predict SH, whereas our LDAN only uses one input face image.
Their results are also achieved by testing on the original high resolution face images of the MultiPie data set.
3DMM cannot predict accurate face normals for face images of $64\times 64$, leading to inaccurate lighting estimations for \cite{Forgery17} on such low resolution images\footnote{Personal communication with B. Peng, co-author of \cite{Forgery17}.}.
We compare the proposed method with their method (denoted as `3D lighting') in Figure~\ref{fig:ROC}. 
Following the same setup of the experiment on MultiPie, $20,000$ pairs of face images are generated using the data provided by \cite{Forgery17}.
Half of these face image pairs are taken under the same lighting condition and the other half taken under different lighting conditions. 
We carry out the verification study on these image pairs.
Euclidean distance and the measurement provided by \cite{Forgery07} (denoted as $Q$-measure) are applied for testing and the ROC curves are shown in Figure~\ref{fig:ROC} (a) and (b) respectively.
We notice that using Euclidean distance, LDAN outperforms `3D lighting', however under the $Q$-measure, `3D lighting' performs much better.
%
This $Q$-measure\footnote{See supplementary materials for details of $Q$-measure.} ignores the DC component of SH for image forgery purposes; though `3D lighting' can predict first and second order components of SH accurately, it could not predict an accurate DC component.
Actually, some of the DC components of SH predicted by `3D lighting' are negative, which is impossible in reality.
While DC components may not be important for image forgery detection, it is crucial for a useful lighting representation, especially for application such as image editing.
We also notice that under both measurements, LDAN outperforms SIRFS SH, which further confirms the effectiveness of the proposed method.

\subsection{Running Time}
\label{sec:RunningTime}
We run experiments on a workstation with $4$ Intel Xeon CPUs and $80$ GB memories.
While running on GPU, we use one NVIDIA GeForce TITAN X.
For a $64\times 64$ RGB face image, SIRFS \cite{SIRFS} takes $47$ second to predict the lighting parameters.
The proposed deep CNN can predict $390$ such face images on CPU and $2400$ face images on GPU per second, so it is potentially $100,000$ times faster.
%

\section{Conclusion}
\label{sec:conclusion}
In this paper, we propose a lighting regression network to predict Spherical Harmonics of environment lighting from face images. 
Lacking the ground truth labels for real face images, we applied an existing method to get noisy ground truth.
To alleviate the effect of noise, we propose to apply the idea of adversarial networks and use synthetic face images with known ground truth to help train a deep CNN for lighting regression.
Compared with existing methods, the proposed method is more efficient and the experimental results show it improves the performance significantly.
%

\begingroup
\setstretch{1}
\setlength{\bibsep}{1pt}
\bibliographystyle{plain}
{\small
\bibliography{egbib}}
\endgroup

\clearpage

\section{Supplementary Material}
\subsection{Spherical Harmonics}
The 9 dimensional spherical harmonics in terms of Cartesian coordinates of the surface normal $\vec{n}=(x,y,z)$ is:
\begin{eqnarray}
Y_{00} = \frac{1}{\sqrt{4\pi}} \qquad  \qquad  Y_{10} = \sqrt{\frac{3}{4\pi}}z \qquad \qquad Y_{11}^{e}=\sqrt{\frac{3}{4\pi}}x& \nonumber \\
Y_{11}^{o} = \sqrt{\frac{3}{4\pi}}y   \qquad Y_{20} = \frac{1}{2}\sqrt{\frac{5}{4\pi}}(3z^2-1) \qquad    Y_{21}^{e}=3\sqrt{\frac{5}{12\pi}}xz  \\
Y_{21}^{o} = 3\sqrt{\frac{5}{12\pi}}yz   \qquad Y_{22}^{e} = \frac{3}{2}\sqrt{\frac{5}{12\pi}}(x^2-y^2) \qquad    Y_{22}^{0}=3\sqrt{\frac{5}{12\pi}}xy \nonumber
\end{eqnarray}

\subsection{Details of Model B and Model C}
We show the details of Model B and Model C in our paper.
Figure~\ref{fig:Model_BC} (a) and (b) illustrates the structure of Model B and Model C respectively.
The objective function of training Model B and Model C is the same with that of training LDAN, shown in Equation~(\ref{eq:wgan_1}).
\begin{eqnarray}
\min_{\mathcal{R}, \mathcal{S}, \mathcal{L}}\max_{\mathcal{D}} 
& \underbrace{\mathbb{E}_{\mathcal{S}(\mathbf{s})\sim \mathbb{P}_s }[\mathcal{D}(\mathcal{S}(\mathbf{s})] - \mathbb{E}_{\mathcal{R}(\mathbf{r})\sim \mathbb{P}_r }[\mathcal{D}(\mathcal{R}(\mathbf{r}))]}_\text{GAN loss} 
+ \underbrace{\mu \sum_i(\mathcal{L}(\mathcal{R}(\mathbf{r}_{i})) - \mathbf{\hat{y}}_{ri})^2}_\text{regression loss for real} \nonumber \\
&+ \sum_{(i,j)\in \Omega}(
\underbrace{\nu [(\mathcal{L}(\mathcal{S}(\mathbf{s}_{i})) - \mathbf{y}_{si}^{*})^2 + (\mathcal{L}(\mathcal{S}(\mathbf{s}_{j})) - \mathbf{y}_{si}^{*})^2]}_\text{regression loss for synthetic} + 
\underbrace{\lambda(\mathcal{S}(\mathbf{s}_{i}) - \mathcal{S}(\mathbf{s}_{j}))^2}_\text{feature loss}) \label{eq:wgan_1}
\end{eqnarray}

The difference of these two models and LDAN is that these two models try to drag the distribution of lighting related features of real and synthetic data towards each other, while LDAN push the distribution of lighting related features of real data towards that of synthetic data.
This difference is illustrated in Figure~\ref{fig:idea}.
Model B is inspired by \cite{DomainGan_ICML2015} and real and synthetic data share the same feature net.
As a result, for Model B, $\mathcal{R}$ and $\mathcal{S}$ are the same.
Model C, on the other hand, is inspired by \cite{DomainGan_Kuniaki}, and defines two different feature nets for real and synthetic data.

Similar to LDAN, the lighting net in Model B and Model C is trained only using synthetic data so that it will not be affected by the noise in the labels of real data.
Algorithm~\ref{al:CNN_AB} shows the details of how to train Model B and Model C.

\begin{algorithm}[H]
\caption{Training procedure for Model B and C}
\label{al:CNN_AB}
\begin{algorithmic}[1]
\For {number of training iterations}
	\For {$k$= 1 to 5 epochs}
    	\State{Train discriminator $\mathcal{D}$ through the following loss using RMSProp\cite{RMSProp}:
        \begin{eqnarray}
        	\max_{\mathcal{D}}\mathbb{E}_{\mathcal{S}(\mathbf{s})\sim \mathbb{P}_s }[\mathcal{D}(\mathcal{S}(\mathbf{s}))] - \mathbb{E}_{\mathcal{R}(\mathbf{r})\sim \mathbb{P}_r}[\mathcal{D}(\mathcal{R}(\mathbf{r}))] \nonumber
        \end{eqnarray}} 
    \EndFor
     \For {$k$=1 to 2 epochs}
    	\State {Train $\mathcal{L}$ through the following loss using Adadelta \cite{Adadelta}:
        \begin{equation}
        	\min_{\mathcal{L}} \sum_{(i,j)\in \Omega} \nu [(\mathcal{L}( \mathcal{S}(\mathbf{s}_{i})) - \mathbf{y}_{si}^{*})^2 + (\mathcal{L}(\mathcal{R}(\mathbf{s}_{j})) - \mathbf{y}_{si}^{*})^2] \nonumber
        \end{equation}}
    \EndFor
    \For {$k$=1 to 2 epochs}
    	\State {Train $\mathcal{R}$ and $\mathcal{S}$ using Equation~\ref{eq:wgan_1} by Adadelta.}
    \EndFor
\EndFor
\end{algorithmic}
\end{algorithm}

\begin{figure}
\begin{adjustbox}{center}
\begin{tabular}{c}
\includegraphics[width=10cm]{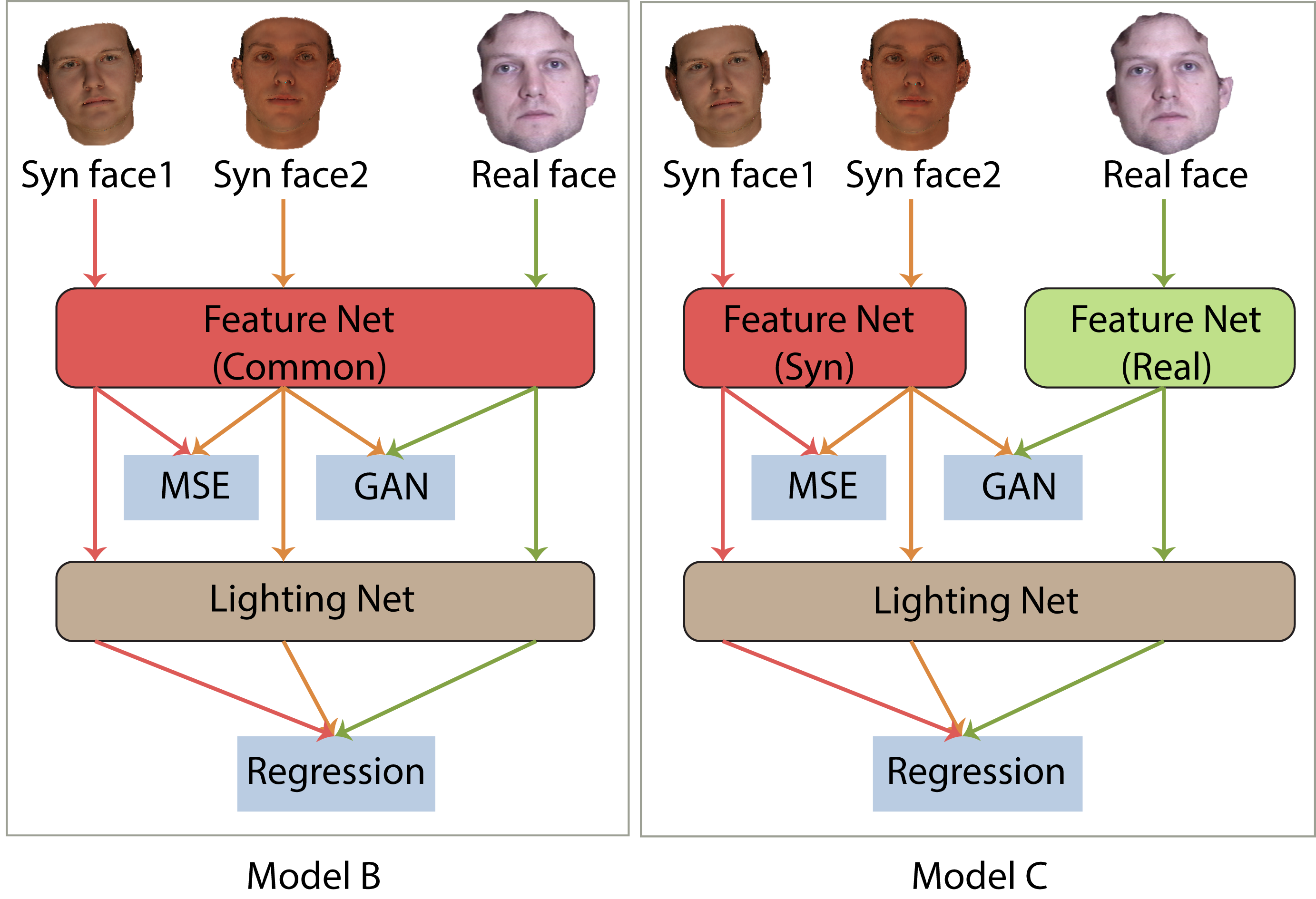} 
\end{tabular}
\end{adjustbox}
\caption{Model B and Model C mentioned in the paper.}
\label{fig:Model_BC}
\end{figure}

\begin{figure}
\begin{adjustbox}{center}
\begin{tabular}{cc}
\includegraphics[width=6cm]{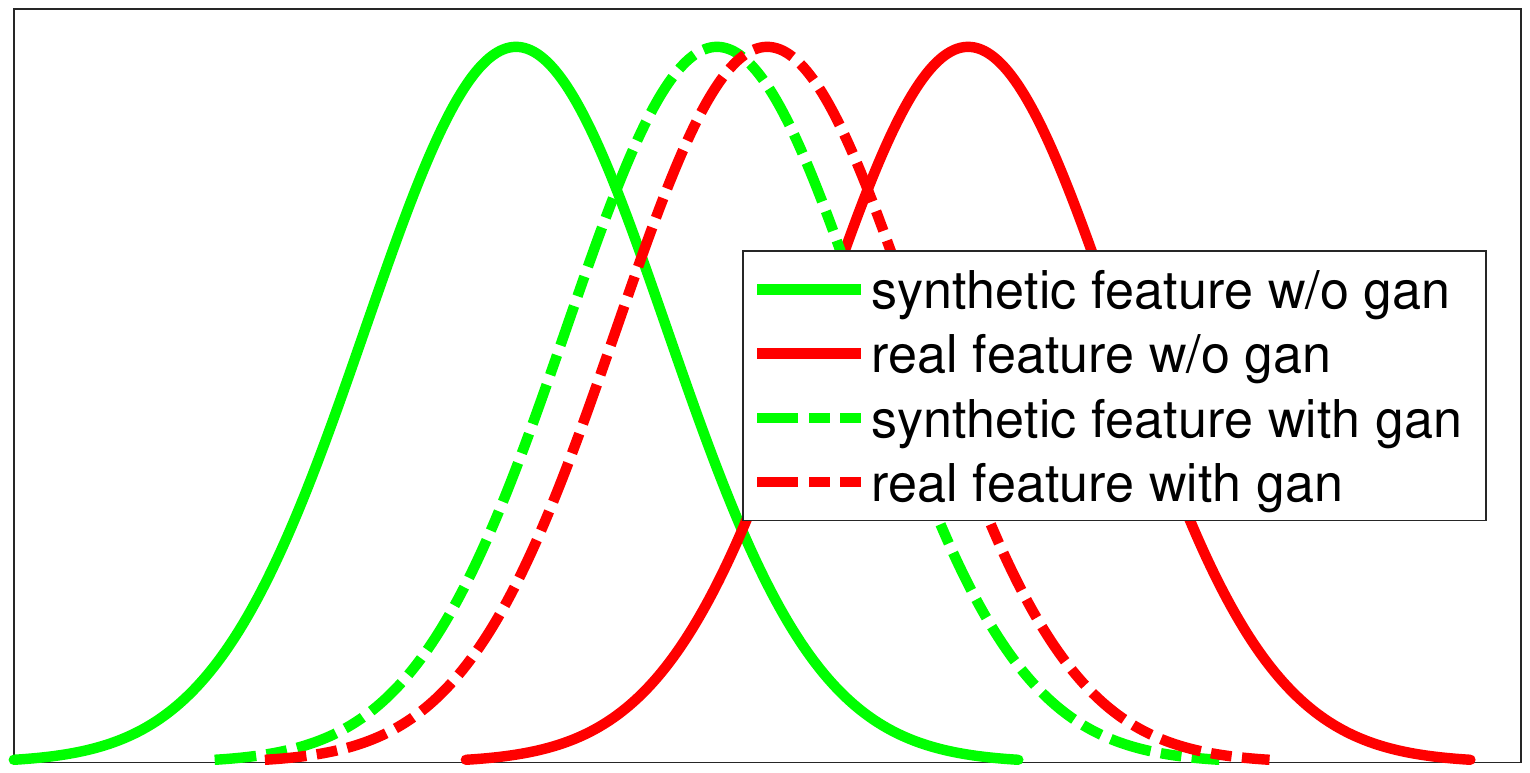}&
\includegraphics[width=6cm]{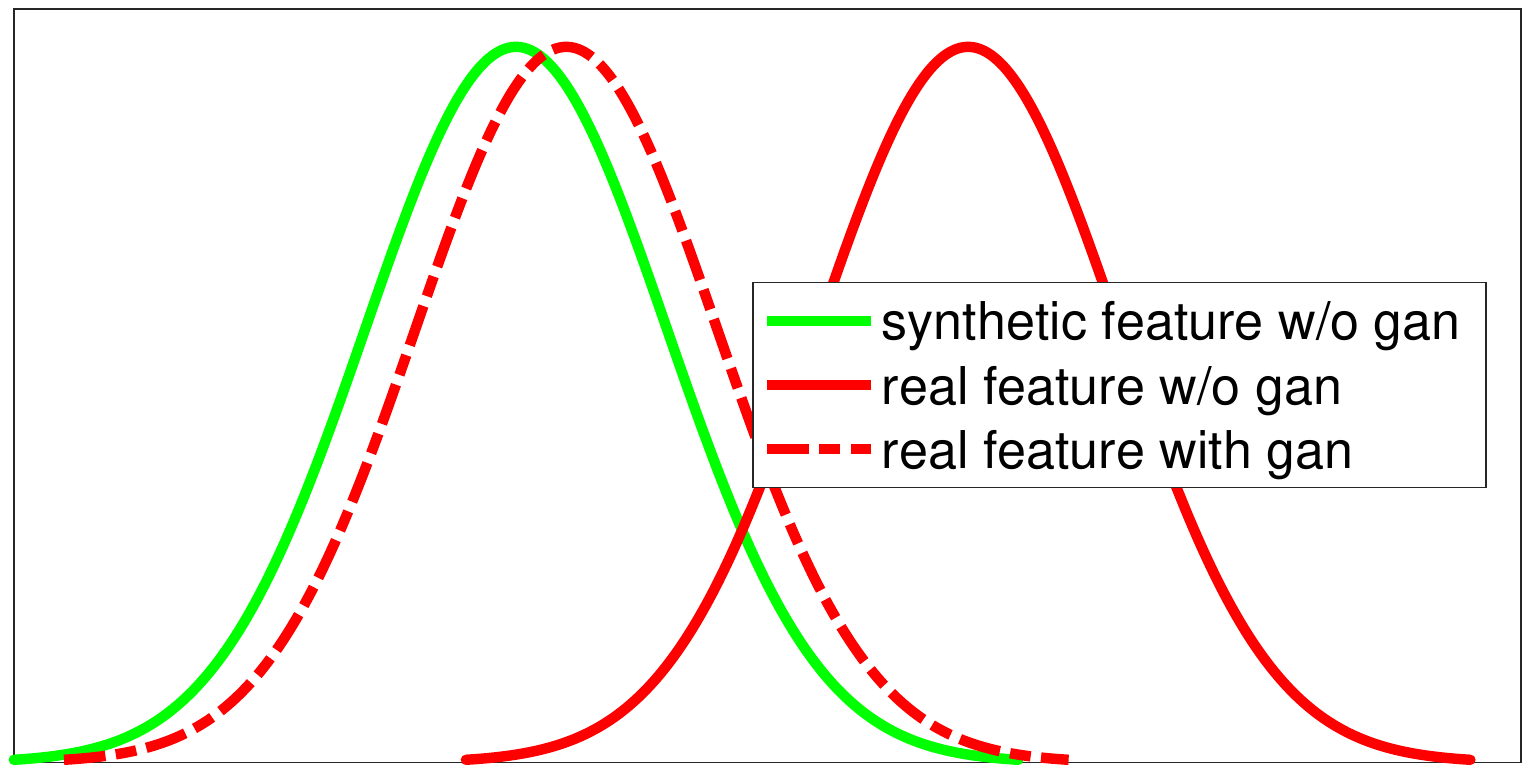}\\
(a) Model B and C & (b) LDAN
\end{tabular}
\end{adjustbox}
\caption{(a) illustrates the distribution of lighting related features of real and synthetic data before and after applying GAN loss for Model B and Model C. (b) illustrates the distribution of lighting related features of real and synthetic data before and after applying GAN loss for LDAN.}
\label{fig:idea}
\end{figure}

\subsection{Structure of Networks}

We show the structure of our networks in this section. 
As discussed in the paper, we apply the structure of ResNet \cite{ResNet} to define our feature net.
Figure~\ref{fig:network} (a) shows the structure of the feature net.
A block like ``3$\times 3$ conv 16'' means a convolutional layer with 16 filters, the size of each filter is $3 \times 3 \times n$ where $n$ is the number of input channels. 
This convolutional layer is followed by a batch normalization layer and a ReLU layer.
A block like ``3 $\times$ 3 conv 32, /2'' has similar meaning, the difference is that the stride of the convolution is $2$ so it down samples the data by a factor of $2$.
The output of the feature net is a $128$ dimensional feature.

Figure~\ref{fig:network} (b) shows the structure of the lighting net.
``FC ReLU 128'' means a fully connected layer whose number of outputs is $128$ followed by a ReLU layer.
``Dropput'' means a dropout layer with dropout ratio being $0.5$.
``FC, 18'' means a fully connected layer with $18$ outputs.

Figure~\ref{fig:network} (c) shows the structure of the discriminator.
``FC tanh, 1'' means a fully connected layer with just 1 output followed by a tanh layer. 
The meaning of the rest of the blocks is the same to those of lighting net.

\begin{figure}[H]
\begin{adjustbox}{center}
\begin{tabular}{ccc}
\includegraphics[width=5.5cm]{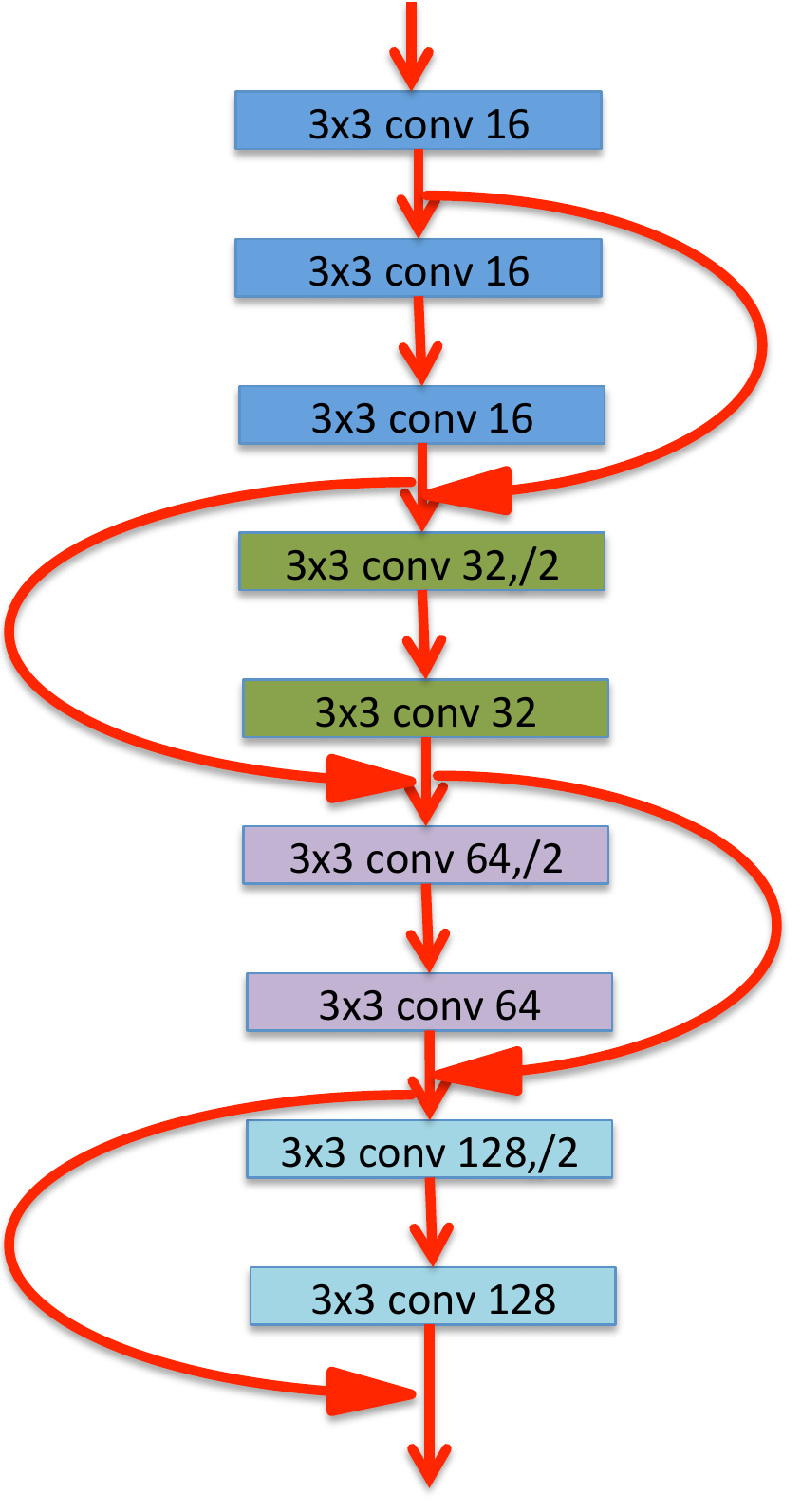}&
\includegraphics[width=3.2cm]{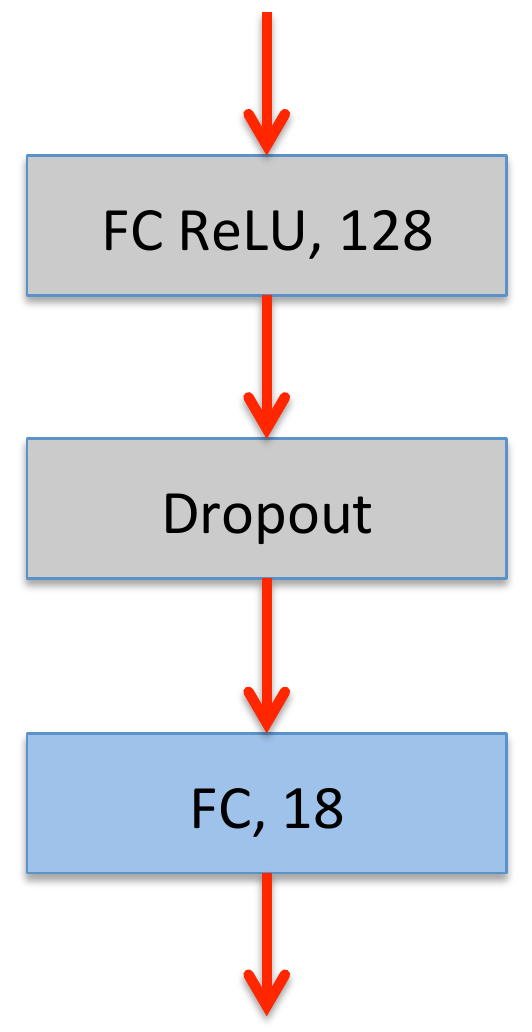}&
\includegraphics[width=3cm]{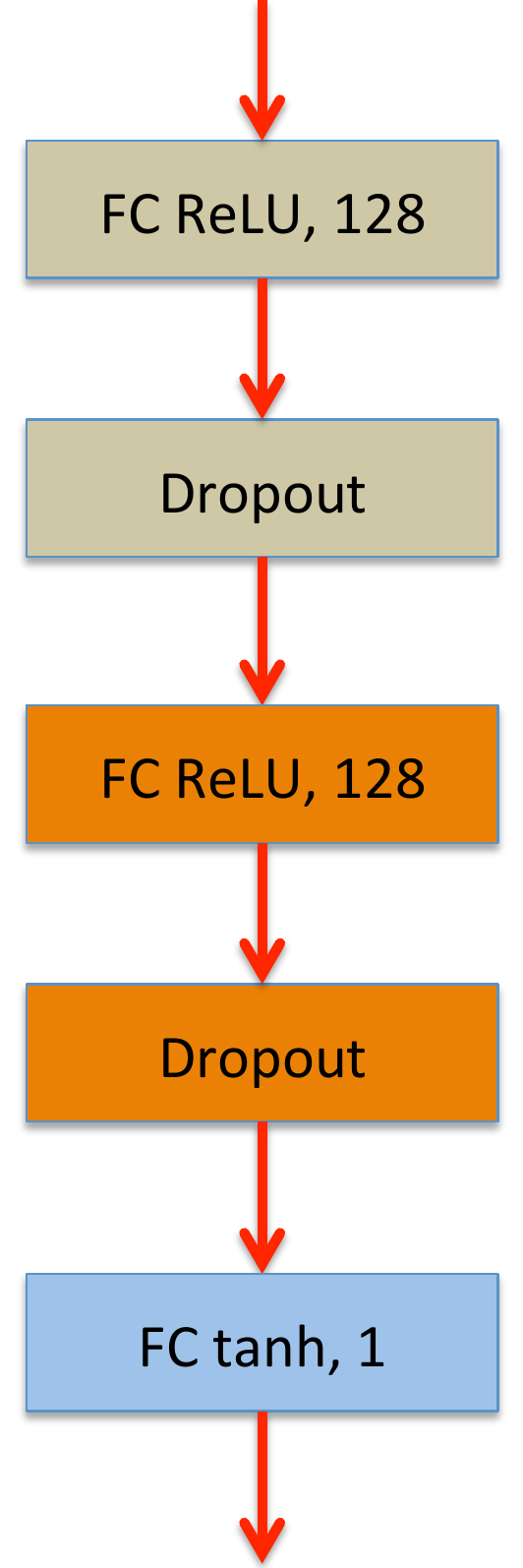}\\
(a) Feature net & (b) Lighting net & (c) Discriminator
\end{tabular}
\end{adjustbox}
\caption{(a), (b) and (c) show the structure of feature net, lighting net and discriminator used in our paper.}
\label{fig:network}
\end{figure}

\subsection{Definition of Q-measure}
Let $\mathbf{y}_1$ and $\mathbf{y}_2$ be two Spherical Harmonics (SH). 
Using these two SH to render two hemispheres and arrange the pixels in to two vectors, we get:
\begin{eqnarray}
\mathbf{x}_1 &= Y\mathbf{y}_1 \nonumber \\
\mathbf{x}_2 &= Y\mathbf{y}_2 ,
\end{eqnarray}
where Y is a $n \times 9$ matrix and $n$ is number of pixels. Each column of $Y$ corresponds to one SH base image.
Let $\mathbf{\hat{x}}_1$ and $\mathbf{\hat{x}}_2$ be the vector of $\mathbf{x}_1$ and $\mathbf{x}_2$ subtracting the mean of their elements:
\begin{eqnarray}
\mathbf{\hat{x}}_1 &=& \mathbf{x}_1 - \text{mean}(\mathbf{x}_1) \nonumber \\
\mathbf{\hat{x}}_2 &=& \mathbf{x}_2 - \text{mean}(\mathbf{x}_2) 
\end{eqnarray}

The correlation of $\mathbf{\hat{x}}_1$ and $\mathbf{\hat{x}}_2$ can be computed as:
\begin{eqnarray}
\text{corr}(\mathbf{\hat{x}}_1, \mathbf{\hat{x}}_2) &=& \frac{\mathbf{\hat{x}}_1^T\mathbf{\hat{x}}_2}{||\mathbf{\hat{x}}_1|| ||\mathbf{\hat{x}}_2||} \label{eq:Q_1}\\
&=&\frac{\mathbf{y}_1^TQ\mathbf{y}_2}{\sqrt{\mathbf{y}_1^TQ\mathbf{y}_1} \sqrt{\mathbf{y}_2^TQ\mathbf{y}_2}}. \label{eq:Q_2}
\end{eqnarray}
Please refer to \cite{Forgery07} for the details of how to get Equation~(\ref{eq:Q_2}) from Equation~(\ref{eq:Q_1}).
The $Q$-measure proposed in \cite{Forgery07} is defined as follows:
\begin{eqnarray}
\text{Distance}(\mathbf{y}_1, \mathbf{y}_2) = \frac{1}{2}(1-\frac{\mathbf{y}_1^TQ\mathbf{y}_2}{\sqrt{\mathbf{y}_1^TQ\mathbf{y}_1} \sqrt{\mathbf{y}_2^TQ\mathbf{y}_2}}),
\end{eqnarray}
where
\begin{eqnarray}
Q=\left(\begin{array}{ccccccccc}
0 & 0 & 0 & 0 & 0 & 0 & 0 & 0 & 0\\
0 & \frac{\pi}{36} & 0 & 0 & \frac{\sqrt{5}\pi}{64\sqrt{3}} & 0 & 0 & 0 & 0\\
0 & 0 & \frac{\pi}{9} & 0 & 0 & \frac{\sqrt{5}\pi}{64} & 0 & 0 & 0 \\
0 & 0 & 0 & \frac{\pi}{9} & 0 & 0 & \frac{\sqrt{5}\pi}{64} & 0 & 0 \\
0 & \frac{\sqrt{5}\pi}{64\sqrt{3}} & 0 & 0 & \frac{\pi}{64} & 0 & 0 & 0 & 0 \\
0 & 0 & \frac{\sqrt{5}\pi}{64} & 0 & 0 & \frac{\pi}{64} & 0 & 0 & 0 \\
0 & 0 & 0 & \frac{\sqrt{5}\pi}{64} & 0 & 0 & \frac{\pi}{64} & 0& 0\\
0 & 0 & 0 & 0 & 0 & 0 & 0 &\frac{\pi}{64} & 0\\
0 & 0 & 0 & 0 & 0 & 0 & 0 & 0 &\frac{\pi}{64}
\end{array}
\right)
\end{eqnarray}
Note the definition of $Q$ is a little different from \cite{Forgery07} due to the difference of arrangement of the elements in SH.

\clearpage
\subsection{More Visual Results}
We show more synthetic faces rendered using SH estimated by the proposed model in Figure~\ref{fig:Results}.
\begin{figure}[H]
\begin{adjustbox}{center}
\begin{tabular}{ccc}
\includegraphics[width=5cm]{new_lighting_example/001-01-01-051-01.png}&
\includegraphics[width=5cm]{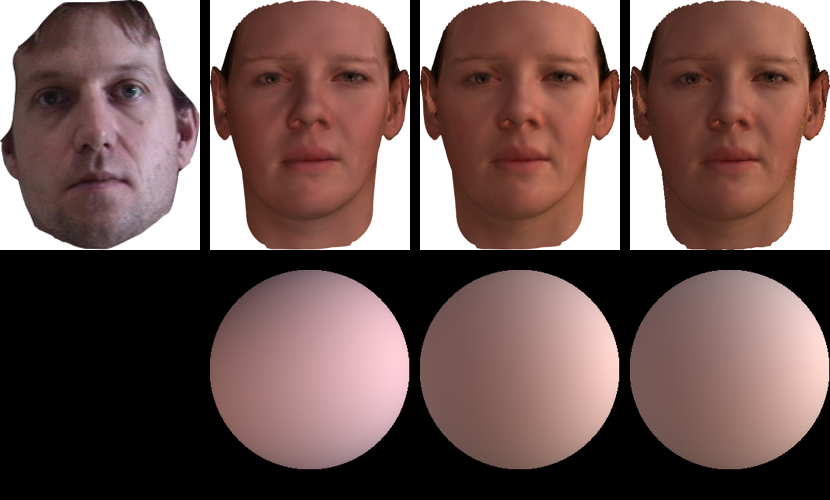}&
\includegraphics[width=5cm]{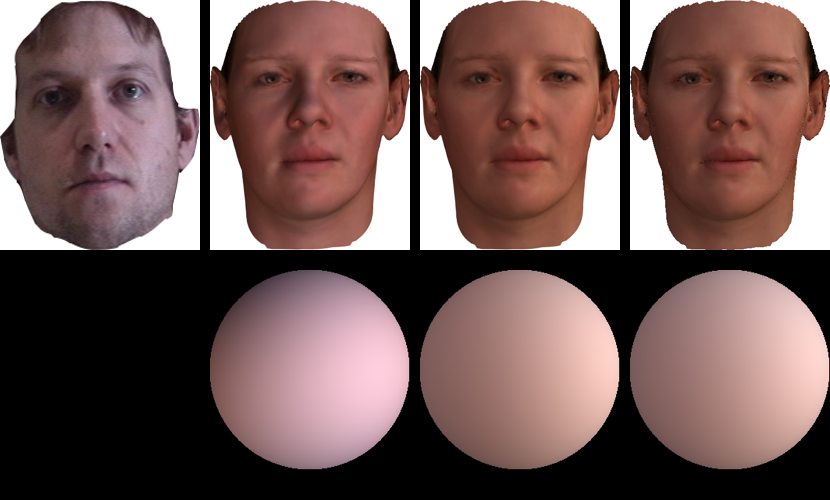}\\
\includegraphics[width=5cm]{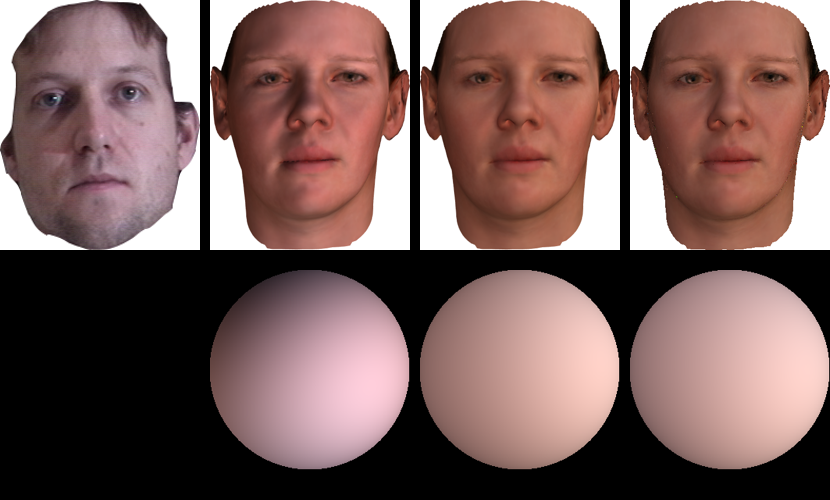}&
\includegraphics[width=5cm]{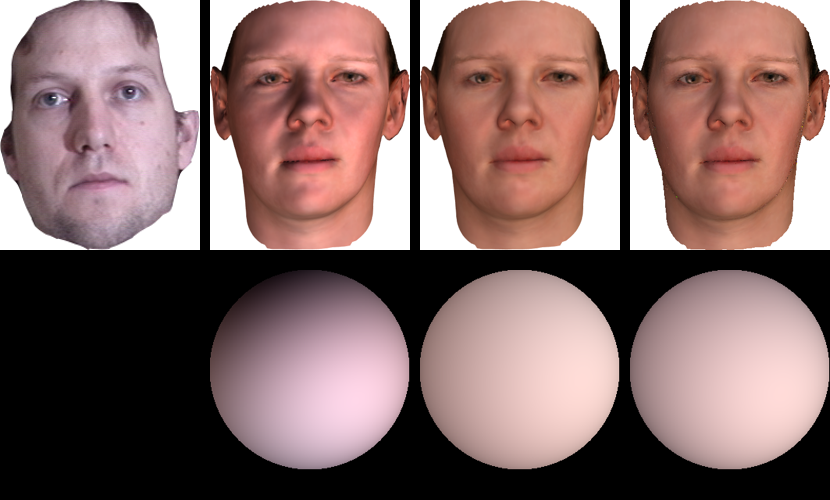}&
\includegraphics[width=5cm]{new_lighting_example/001-01-01-051-06.png}\\
\includegraphics[width=5cm]{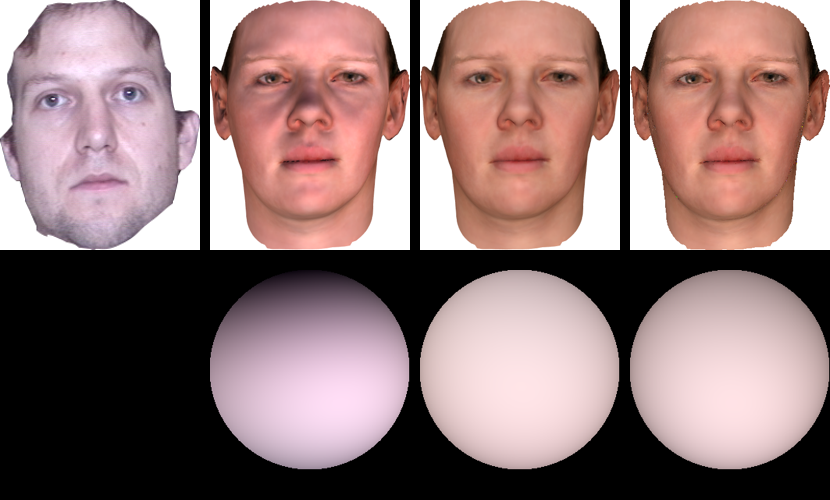}&
\includegraphics[width=5cm]{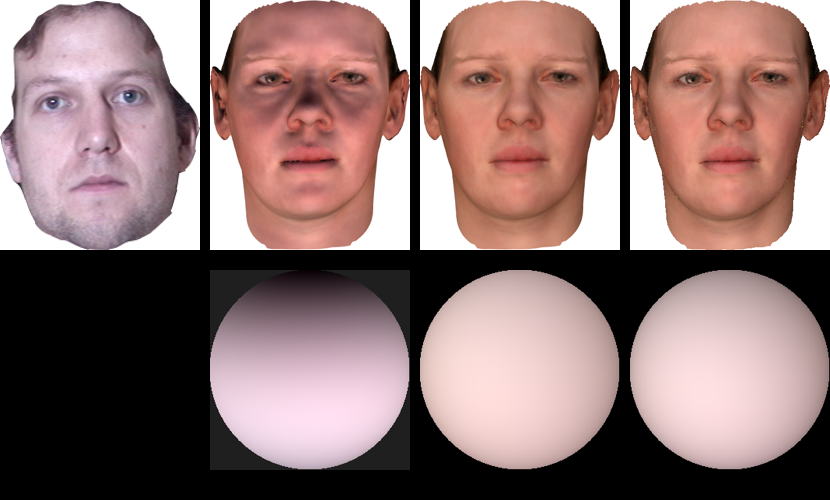}&
\includegraphics[width=5cm]{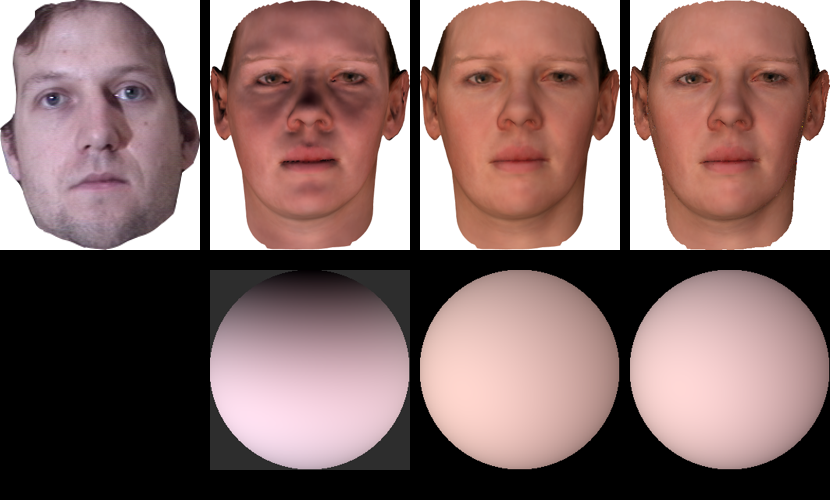}\\
\includegraphics[width=5cm]{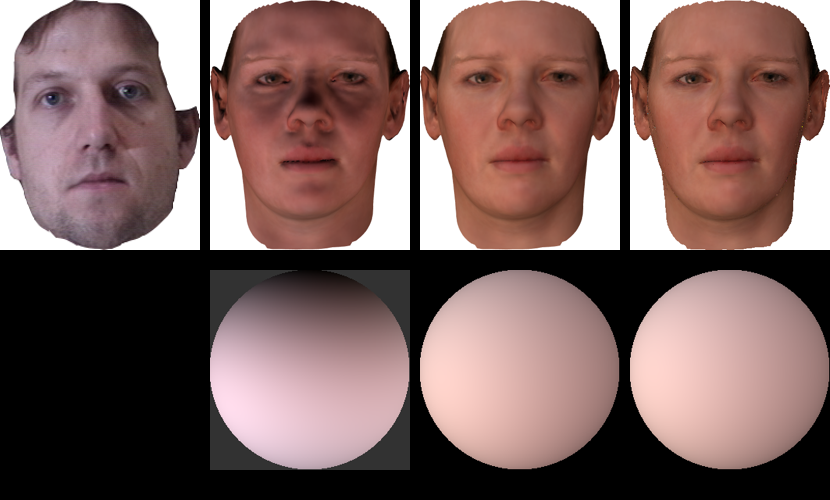}&
\includegraphics[width=5cm]{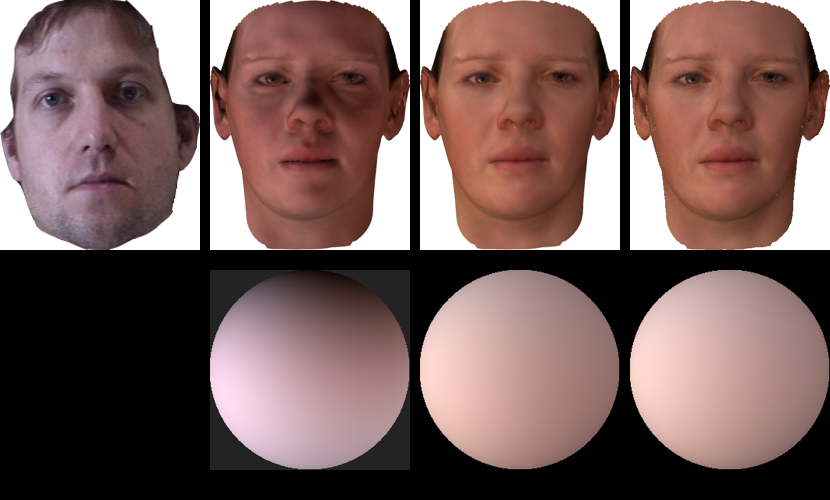}&
\includegraphics[width=5cm]{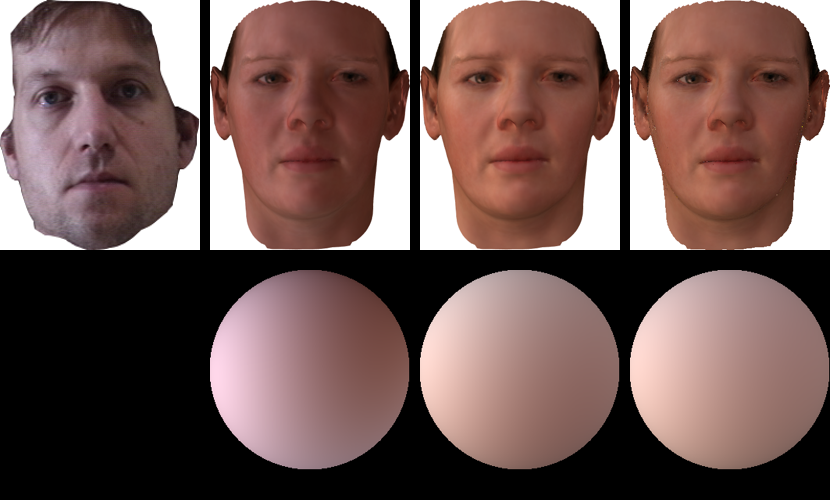}\\
\includegraphics[width=5cm]{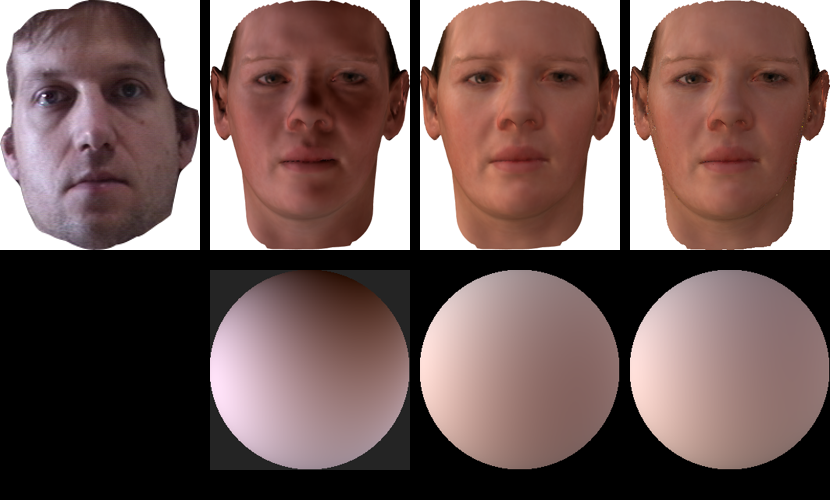}&
\includegraphics[width=5cm]{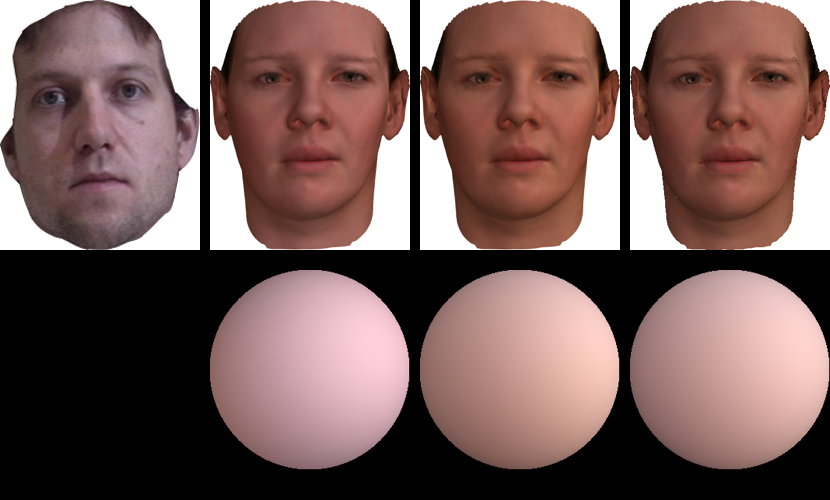}&
\includegraphics[width=5cm]{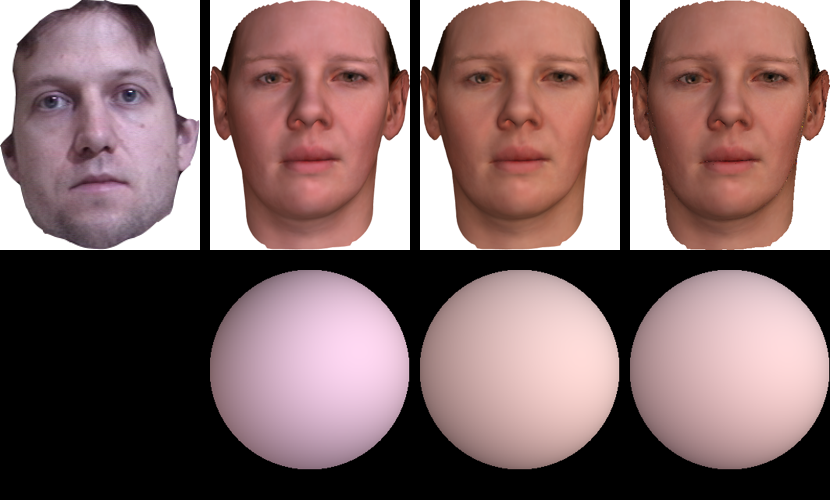}\\
\includegraphics[width=5cm]{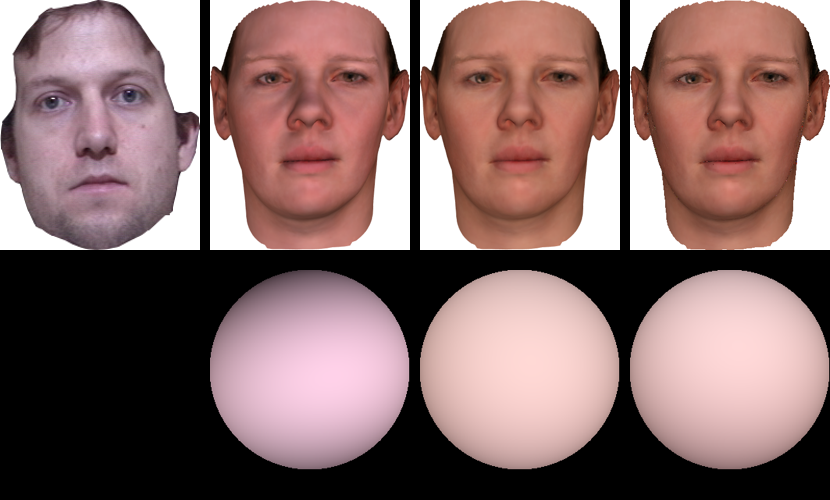}&
\includegraphics[width=5cm]{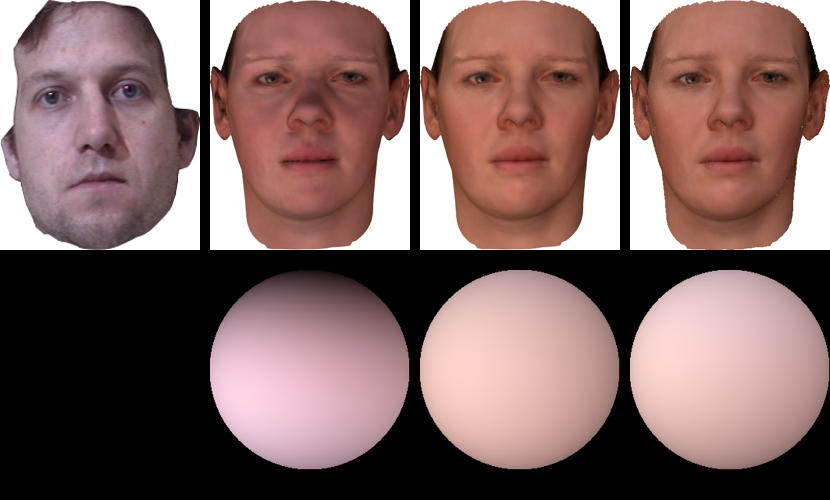}&
\includegraphics[width=5cm]{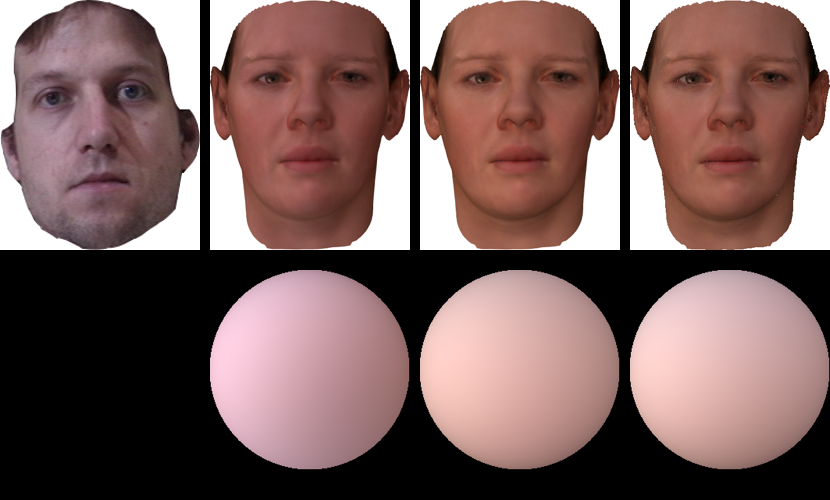}
\end{tabular}
\end{adjustbox}
\label{fig:Results}
\caption{The images in the first row of are: MultiPie face image, synthetic face image rendered using lighting estimated by SIRFS SH, REAL and LDAN from the MultiPie face image. The second row shows a hemisphere rendered by the corresponding lighting.}
\end{figure}

\end{document}